# CIMLA: Interpretable AI for inference of differential causal networks

Payam Dibaeinia[1], Saurabh Sinha[2*]

[1] Department of Computer Science, University of Illinois at Urbana-Champaign, Urbana, IL, 61801, USA
[2] Wallace H. Coulter Department of Biomedical Engineering, Georgia Institute of Technology, Atlanta, GA, 30332, USA

[*] Correspondence: saurabh.sinha@bme.gatech.edu

## Abstract

The discovery of causal relationships from high-dimensional data is a major open problem in bioinformatics. Machine learning and feature attribution models have shown great promise in this context but lack causal interpretation. Here, we show that a popular feature attribution model estimates a causal quantity reflecting the influence of one variable on another, under certain assumptions. We leverage this insight to implement a new tool, CIMLA, for discovering condition-dependent changes in causal relationships. We then use CIMLA to identify differences in gene regulatory networks between biological conditions, a problem that has received great attention in recent years. Using extensive benchmarking on simulated data sets, we show that CIMLA is more robust to confounding variables and is more accurate than leading methods. Finally, we employ CIMLA to analyze a previously published single-cell RNA-seq data set collected from subjects with and without Alzheimer's disease (AD), discovering several potential regulators of AD.

# INTRODUCTION

A key challenge in bioinformatics today is to extract causal relationships from omics data. A common approach involves calculating statistical associations between pairs of variables, such as between expression levels of two genes[1], between alleles and phenotype[2] or gene expression[3], etc. Pairwise associations are error-prone due to the presence of confounding variables[4]. This has led to multivariable regression and machine learning (ML) models that learn a mapping between the target variable (e.g., gene expression) and multiple potentially causal variables or "features" (e.g., expression levels of transcription factors)[5–8]. After model training, various approaches are used to extract the importance of each feature to the model[6,8,9], a process informally known as "interpretation" of the model. Model interpretation is straight-forward in case of linear models, where the coefficient of each feature serves as its importance score, but is challenging for non-linear ML models (e.g., Random Forests and Convolutional Neural Networks) with intense on-going efforts towards improvement of "local"[10–13] (for each sample) and "global"[14–17] (across all samples) feature attribution models.

The ultimate goal of model interpretation in bioinformatics is to infer "causal" relationships that can provide biological hypotheses testable via experimental perturbation. Indeed, the search for causal relationships is a hallmark of much of natural and social sciences, and successful paradigms include interventions in randomized control trials (RCT)[18] and temporal data modeling[19]. However, causality is not explicitly addressed in the above-mentioned approaches to ML model interpretation. ML models themselves are already recognized as tools for counterfactual reasoning [20,21], but the feature attribution models used with them lack a formal causal interpretation, focusing instead on quantifying the importance of a variable to the ML model's output[11,22,23]. We believe this is a significant conceptual gap in the modern "interpretable AI" movement and has led to confusion about the relative strengths and weaknesses of different feature attribution models[24]. Our work is an attempt to bridge this gap and use the resulting insights to address an important bioinformatics goal.

In recent times, a popular formalism for causal relationships and counterfactual inference with observational data has emerged in the form of Pearl's do-calculus[25]. It provides a mathematical framework to clearly describe causal relationships and, when possible, infer their strengths from data. This is referred to as calculating the value of a "causal estimand" from data. On the other hand, a recent breakthrough in model interpretation is the "SHapley Additive exPlanations (SHAP)" feature attribution model[23], which adopts a game theoretic approach to provide a measure of local feature importance in an ML model. SHAP unifies and improves on an important group of existing feature attribution models. Our main theoretical contribution is to show that the SHAP score estimates a causal quantity defined in Pearl's causal inference framework. We define this causal estimand and present the conditions under which it is estimated by the SHAP score.

Building on the newfound intuition about the SHAP score as a measure of causal influence, we propose a framework to quantify how much a feature's influence on a target variable *differs* between two data sets. The causal interpretation of SHAP scores is critical here, as it enables us to directly compare these scores between models trained on different data sets. We present a tool called CIMLA (Counterfactual Inference by Machine Learning and Attribution Models) that implements this framework for inferring differential associations. CIMLA can be a powerful tool for a variety of bioinformatics challenges where one seeks to identify biological relationships that have changed between conditions (such as disease and healthy, treatment and control, etc.). To demonstrate this, we use the CIMLA score to recover regulatory relationships that differ between transcriptomics data sets from two conditions, a problem also known as differential gene regulatory network ("dGRN") reconstruction[26].

The common approach to dGRN reconstruction is based on pairwise co-expression analysis to identify regulatory relationships in each data set, followed by tests of significance of the change in their regulatory strength between the data sets (reviewed by Bhuva et al.[26]). To address the well-known limitations of pairwise association analysis, multivariable linear regression models are sometimes used[27,28], with comparison of corresponding regression coefficients to infer differential regulation. Other authors have explored joint modeling of two data sets in search of shared as well as exclusive regulatory associations[29,30]. All of the above strategies have relied on linear modeling of gene expression as a function of other genes' or transcription factors' expression. Yet, it is known that GRN reconstruction (on a single data set) benefits from incorporating non-linear dependencies, and leading methods rely on ML approaches such as Regression Random Forests (GENIE3[31]), or Gradient Boosting Regression (GRNBoost2[32], ENNET[33]) to achieve cutting-edge performance on standard benchmarks. Thus, it is natural to ask if dGRN reconstruction can be improved by training an ML model on either data set and comparing feature importance scores between the two models. This is exactly the task that CIMLA is designed to perform, making it an ideal test case for the new measure of differential feature importance.

To summarize, we make the following main contributions in this work. First, we define a quantity that measures the causal influence of a variable on an outcome, in the presence of other causal covariates and confounders, and show that under certain assumptions this quantity is estimated by the SHAP score of Lundberg et al.[10,23]. Second, we present the CIMLA score to measure changes in feature importance between two observational data sets. It is based on the SHAP score and thus inherits its desirable feature attribution properties, while also having an intuitive causal interpretation for changes in influence. Third, we systematically assess CIMLA for the task of dGRN inference, using synthetic benchmarks based on state-of-the-art expression simulators, and demonstrate its advantages over existing methods, especially in scenarios where conditional differences in GRN are confounded by other changes. Finally, we employ CIMLA to study the dGRN underlying Alzheimer's disease (AD), using one of the largest published single-nucleus RNA-seq (snRNA-seq) datasets for AD, and find two known regulators of AD – CREB3 and NEUROD6 – to undergo substantial changes in their respective regulons.

## RESULTS

### Overview of CIMLA

Our goal is to identify features (variables) whose association with a measurable trait changes between two populations, using observational data. The two populations may differ in terms of some condition such as disease treatment status, and we seek to quantify the impact of this varying condition on the feature-trait associations. For instance, in the context of dGRN inference the features are expressions of a set of transcription factors (Figure 1A, $X = \{X_i\}_{i \in \{1..m\}}$) and the trait of interest is a target gene's expression (Figure 1A, $Y$). The observational data comprise expression profiles ($\{X_i\}_{i \in \{1..m\}}, Y$) of multiple samples in each condition, and the task is to detect if an association between say $X_t$ and $Y$ changes between the two populations. We call such an association a "differential association". There are likely to be unobserved confounders (Figure 1A, $Z'$) that may lead to spurious detection of differential associations. CIMLA approaches differential association detection from a causal inference perspective, for enhanced robustness toward confounders. We present here a procedural overview of CIMLA, with accompanying details included in Methods. The terminology here is tailored to the dGRN inference problem but the procedure applies broadly to other applications of differential association detection.

*Step 1 ("ML module"):* Given transcriptomic data from two populations (Figure 1B), CIMLA first trains an ML model to predict a target gene's expression (i.e., outcome variable) as a function of TFs' expression (covariates), separately for each population (Figure 1C). Formally, let $D_c$ denote the data for the population under condition $C = c$, $c \in \{0,1\}$. This step trains, for each $D_c$, an ML model $f_c$ capable of predicting $Y$ from $X$. The current CIMLA implementation relies on random forest[34] (RF) and Neural Networks (NN) with dropout[35] to better tackle multi-collinearity among covariates. (Gradient Boosting is also supported but it was not tested here.)

*Step 2 ("Interpretation module"):* Next, trained ML models and the data are passed to an interpretation module to assess the contribution of each covariate to the predicted outcome under the two ML models, locally at each sample (Figure 1D). Formally, it calculates, for either model $f_c$ ($c \in \{0,1\}$), the SHAP value[10,23] $\phi_t(f_c, x)$ that quantifies the contribution of covariate $X_t$ to the model's output at $X = x$. This step relies on TreeSHAP[10] and DeepSHAP[23] to estimate local contributions for RF and NN models respectively. This step is where CIMLA adopts a causal inference framework, as we show that the SHAP value $\phi_t(f_c, x)$ approximates a well-defined causal quantity reflecting the influence of covariate $X_t$ on outcome $Y$ for sample $x$, under a reasonable set of assumptions (see Methods).

*Step 3 ("Aggregation module"):* Finally, the difference in local contribution scores for the two models is computed for each sample and aggregated across all samples in $D_{c=1}$ using a "root mean square" function (Figure 1E). Specifically, for each sample $x$ in $D_{c=1}$, we calculate $\Delta_t(x) = \phi_t(f_{c=1}, x) - \phi_t(f_{c=0}, x)$, and then compute the aggregate $\Lambda_t = \sqrt{E_x \Delta_t(x)^2}$. This aggregate ($\Lambda_t$) is called the "CIMLA score" of covariate $X_t$, and represents the extent to which a causal association (if any) between $X_t$ and outcome $Y$ has changed between the two populations.

The CIMLA implementation has separate modules for the three steps outlined above, allowing for future developers to modify or improve each step independently.

**Benchmarking on simulated transcriptomics data**

As there are no gold standard datasets for evaluating dGRN inference methods, we relied on synthetic data to benchmark CIMLA and compare it with existing methods. We used SERGIO[36] to simulate single-cell expression datasets representing two distinct conditions, arbitrarily termed "case" (C=1) and "control" (C=0), with similar but non-identical GRNs underlying the two datasets. SERGIO is a single-cell expression simulator that synthesizes transcriptomics profiles according to a provided GRN, modeling transcriptional regulation through stochastic differential equations similar to those employed in GeneNetWeaver[37,38]. We used a "reference GRN" previously reported for yeast (see Methods) to construct the case and control GRNs, each containing a subset of the edges in the reference GRN such that the extent of difference between the two GRNs is controlled. The two GRNs were then used by SERGIO for generating case and control expression datasets respectively, with pre-defined profiles of master regulators ("MRs" – TFs that are not targeted by other regulators in the GRN). We repeated this process 15 times to obtain as many GRN pairs, each pair sharing 43-94% of edges (Figure 2A), and corresponding pairs of case and control datasets, each dataset comprising a matrix of expression values with rows representing genes and columns representing cells. The two expression datasets can be provided to a dGRN inference method and since the differences between GRNs are known, it is possible to precisely evaluate its predictions.

For each of the 15 tests, we used CIMLA and other methods to infer dGRN edges. In the results presented in this section, we used RF as the underlying ML model and the TreeSHAP[10] method for interpretation. (We later show the performance of CIMLA with NN models.) For comparison, we tested six co-expression-based methods for dGRN inference (henceforth called "co-expression methods") that were found by Bhuva et al.[26] to be the leaders among all evaluated methods in recovering differential associations. These methods quantify pairwise associations between genes and test for statistical significance of the changes in association metrics between the two datasets in the test. Four of the selected methods rely on correlation metrics (z-score-Pearson, z-score-Spearman[39], MAGIC[40], and DICER[41]), one method uses entropy (entropy method[42]) and the other employs F-statistics (ECF method[43]) to measure pairwise associations. Each method, including CIMLA, is capable of reporting a sorted list of dGRN edges that can be evaluated using standard metrics of classifier performance given a ground truth list of dGRN edges.

We performed our evaluations in two settings – "low-confounding" and "high-confounding" (Figures 2B,C). In the low-confounding setting, we used comparable Master Regulator (MR) expression profiles in generating transcriptomic data from the case and control GRNs; thus, the only difference between the groups was due to differences in their GRNs. On the other hand, for the high-confounding setting (Figure 2C), we set the MR expression profiles to be different in the two groups, mimicking the presence of an unobserved confounder that creates backdoor paths of associations (via $Z'$ in Figure 2C) between population-specific condition ($C$) and target genes ($g$). Figures 2D,E compare the performance of the evaluated methods in terms of AUROC and normalized AUPRC (AUPRC divided by expected AUPRC of random, class size-aware prediction) in the low-confounding setting. We find CIMLA to be competitive with the other methods. (The ECF method performed poorly and was excluded from further analysis.) This is consistent with our expectation that in the absence of confounders, accurate causal inference can be made from associative quantities such as correlation metrics. The z-score-Spearman ("z-score-S") method, which has the second-best median normalized AUPRC after CIMLA, is highly competitive with CIMLA on each of the 15 tests (Figures 2F,G). Also, the performance of both methods generally improves as the extent of difference between case and control GRNs decreases (Figure 2H).

Next, we repeated the above evaluations in the "high-confounding" setting, where the inter-condition difference in GRNs is confounded by differences in the MR profiles that serve as "inputs" to the GRN in generating expression data (see Methods). In this case, as is evident from Figures 2I,J, the performance of all competing methods dropped compared to the low-confounding setting

(Figures 2D,E), leaving a large gap between these methods and CIMLA. Even the second-best method (based on normalized AUPRC) – z-score-Pearson ("z-score-P") – is greatly outperformed by CIMLA on every individual dataset tested (Figures 2K,L). Notably, CIMLA's performance remains robust to the confounder effect (Figure 2H). This robustness to confounders was one of the main motivations behind the causal framework of CIMLA. As the high-confounding setting represents a more realistic scenario, our benchmarking results indicate the importance of causal approaches for dGRN inference.

While co-expression methods explore linear associations, CIMLA can capture non-linear relationships due to its underlying ML models. This motivated us to next compare CIMLA with methods – GENIE3[31] and BoostDiff[44] – that consider the non-linearities of regulatory relationships. BoostDiff[44] employs differential boosted trees using a novel criterion for growing trees based on the difference of the predictive power of regulators in two conditions. GENIE3[31], a leading GRN (not dGRN) inference tool, uses random forests to model each gene's expression in terms of TFs' expression profiles, and uses importance of TFs in the trained model, assessed via an information theoretic tree-based feature importance metric, to score TF-gene pairs. We adapted GENIE3 for dGRN inference by applying it separately on case and control data and comparing the scores of corresponding edges in the two inferred GRNs to rank dGRN edges ("GENIE3-diff"). As shown in Figures 2M,N, CIMLA convincingly outperforms both BoostDiff and GENIE3-diff in terms of both AUROC and normalized AUPRC.

In summary, benchmarking on synthetic data sets demonstrates the advantage of CIMLA over pairwise association as well as multivariable regression-based, linear as well as non-linear modeling approaches to dGRN inference.

**CIMLA prioritizes causation among correlations**

Our results so far suggest that confounders can significantly hinder current methods for dGRN inference and CIMLA is relatively robust to their effect. Our next analyses probed deeper into the source of CIMLA's empirical advantage. First, we defined "delta-correlation" for a TF-gene pair "$tf, g$", as the absolute difference of their pairwise correlations in case and control groups; this is a simple measure of the "correlation signal" that points to differential regulation. Figure 3A compares these delta-correlation values for all TF-gene pairs that belong to the true dGRNs ("differential pairs") and those that do not ("non-differential pairs"). We noted that in the low-confounding setting the distribution of delta-correlation values is clearly different between the two groups of TF-gene pairs; this enables easy detection of differential pairs. In the high-confounding setting, on the other hand, the two distributions are much more similar. Notably, non-differential TF-gene pairs show higher delta-correlations in the high-confounding setting compared to the low-confounding one. This implies that confounders introduce many non-causal group-specific associations that can cause correlation-based methods to err.

Next, we examined (Figures 3B,C) the scores assigned by CIMLA and a correlation-based method, z-score-S, (both converted to percentile) to the differential TF-gene pairs, along with the delta-correlations of those pairs. We noticed, in the low-confounding setting, that z-score-S is largely driven by delta-correlation, as expected (Figure 3B). On the other hand, CIMLA scores are less tightly determined by delta-correlation, and we noted high scores being assigned to many of the differential pairs with small delta-correlations. In high-confounding settings, CIMLA is even less dictated by the delta correlation signal (compared to the low-confounding setting), resulting in a larger gap between the two methods in their ability to identify differential pairs with small delta-correlations (Figure 3C). This enlarged gap is also apparent when we examine the True Positive Rates (TPR) of either method in the high-confounding setting (Figures 3D,E), and is likely the reason behind the greater precision and recall exhibited by CIMLA compared to z-score-S in this setting (Supplementary Figures S1A,B). Interestingly, when we compare the false positive rate (FPR) of the two methods (Supplementary Figures S1C,D), z-score-S shows significantly larger FPRs compared to CIMLA for detecting the large delta-correlation pairs, suggesting that even in low-confounding settings, co-expression methods may be prone to reporting non-causal correlations (Supplementary Figure S1C).

**Evaluations on noisy simulated data**

The evaluations reported above were performed with "clean" simulated datasets, where SERGIO does not introduce any technical noise to the generated data. However, real single-cell RNA-seq data suffer from significant technical noise, especially "dropout", which incorrectly introduces significant numbers of zero values to the expression matrix. To explore the impact of technical noise we repeated parts of the above evaluations on synthetic single-cell expression datasets with dropout. Specifically, starting with one of the 15 tests from above (with fraction of shared GRN edges = 94%), in the high-confounding setting, we added increasing levels of dropout (10%, 20%, …, 70%), with 5 "replicates" per dropout level. Following recommendations from previous work[36], we imputed missing values (zeros) in the expression matrix using MAGIC[45] (with t=2) prior to applying dGRN methods. Moreover, in addition to RF, we also tested a fully connected neural network (NN) as the underlying ML model and DeepSHAP[23] for SHAP score calculations in CIMLA.

We compared the performance of CIMLA with the co-expression methods as well as BoostDiff and GENIE3-diff at varying dropout levels. As expected, dGRN inference deteriorates as the level of noise increases (Supplementary Figures S2A,B). In terms of both AUROC and AUPRC, the two versions of CIMLA – CIMLA-RF and CIMLA-NN – outperform co-expression methods at almost all levels of dropout; however, GENIE3-diff shows competitive performances in terms of AUPRC. These results were obtained for the task of predicting all differential regulators of all genes, i.e., the entire dGRN. In an alternative evaluation, we calculated the AUROC and AUPRC for each target gene separately and examined the per-gene performance metrics over all genes. (Genes with no differential regulators in the ground truth were excluded.) Figures 4A,B compare all methods by the median per-gene AUROC and (normalized) AUPRC respectively, revealing an even clearer advantage for the two CIMLA versions (also see Supplementary Figures S2C-E). We noted that at the highest level of dropout CIMLA-NN achieves better performances compared to CIMLA-RF which suggests greater tolerance of CIMLA-NN to technical noise in the data.

We observed that the CIMLA-RF and CIMLA-NN scores of candidate TFs for a gene show relatively small correlation with each other for many of the target genes (Figure 4C, e.g., median Pearson's correlation of 0.25 at 70% dropout) and also at a more global level (Figure 4D, Pearson's correlation of 0.30). This suggests some complementarity between the two methods, which may allow a combination of the two methods to yield superior performance compared to either alone. We thus defined two related methods – CIMLA-intersection and CIMLA-union – whose dGRN output is the intersection and union respectively of the top 10% scoring TF-gene pairs of CIMLA-RF and CIMLA-NN. We compared their F1 scores with those of the individual methods and z-score-S (also restricted to top 10% predictions) and found that CIMLA-union outperforms the predictions made by other methods, especially at the higher dropout level that is a more realistic scenario for single cell data (Figure 4E).

**Differential GRN of Alzheimer's disease: a case study**
We used CIMLA for the inference of regulatory changes underlying Alzheimer's disease (AD), analyzing a previously published single-nucleus RNA-seq data set obtained from the prefrontal cortex of individuals with or without AD pathology[46]. We used the expression profiles of 44 individuals (see Methods), spanning 16,004 genes in 30,853 cells from 22 individuals in the AD group and 34,852 cells from 22 individuals in the control group. Missing values in the expression matrices of AD and control groups were separately imputed using MAGIC (with t=2)[45]. GRN reconstruction was focused on a set of 2,652 target genes associated with AD according to the DisGeNET database[47] and an unbiased list of 1,289 transcription factors annotated in the AnimalTFDB database[48]. We employed both CIMLA-RF and CIMLA-NN (separately) to identify differential TF-gene pairs between the two groups.

The first step of CIMLA is to train an ML model (Random Forest or Neural Network) to predict each target gene's expression using all TFs as covariates. We assessed the accuracy of these ML models with train-test splits (Figures 5A,B) and noted 1,803 (resp. 2,079) of the 2,652 genes to be reliably modeled by RF (resp. NN) in both AD and control groups, as suggested by their R-squared, $R^2 \geq 0.5$, on both train and test data. The remaining genes and low variance genes were not analyzed further. We assessed the statistical significance of predicted differential regulators, using a background distribution of CIMLA scores obtained on randomized data where group labels (AD/control) of cells had been shuffled (see Methods), and selecting the highest CIMLA score seen for a gene as the significance threshold. Figure 5C shows the distribution of the number of differential regulators of the reliably modeled genes, as predicted by CIMLA-RF and CIMLA-NN, showing that the former is more conservative in its predictions. We limited the resulting dGRNs to target genes that are differentially expressed between AD and control (Methods) and identified the hub TFs and highly targeted genes (TFs and genes respectively included in greatest number of differential pairs) that may play important roles in AD-related dysregulation (Figures 5D,E and Supplementary Figures S3A,B). A survey of the literature revealed evidence in favor of eight of the top 10 hub TFs and nine of the top 10 highly targeted genes in CIMLA-NN's dGRN to be associated with AD and other neurodegenerative diseases (Supplementary Tables S1,S2).

**CIMLA reveals CREB3 and NEUROD6 as potential key regulators of AD**

The TF-gene pairs identified by CIMLA above are likely to include many indirect regulatory relationships, as mechanistic information such as TF-DNA binding was not included in the reconstruction. Thus, we integrated our results with a published GRN for the human brain (from PsychENCODE project[49]), in order to enrich the dGRN for direct regulatory relationships. This GRN (henceforth called "PsychGRN") is based on multi-omics (DNaseI hypersensitivity, Hi-C, TF motifs, RNA-seq) data on multiple psychiatric disorders. We focused on the 2,055 TF-gene edges shared between PsychGRN and the dGRN derived using the CIMLA-union method, involving 401 TFs and 1,114 genes, which we refer to collectively as the "high-confidence dGRN".

For each TF, we extracted its "regulon" (all predicted targets) in the high-confidence dGRN and assigned it a dGRN and a PsychGRN score based on the average dGRN scores and average PsychGRN weights (see Methods) of the included TF-gene relationships, respectively. Figure 5F shows the top 40 largest regulons in the high-confidence dGRN. We find the TF CREB3 to have the largest regulon, targeting 40 genes (Figure 5G), with a dGRN score that is among the top five. CREB3 (cAMP response element binding protein 3) is involved in regulation of Golgi homeostasis and significantly contributes to Central Nervous System (CNS) function and development [50–52]. It is involved in unfolded protein response to endoplasmic reticulum (ER) stress [53], a response activated in AD[54]. CREB3 is also known to regulate GLUT3, a neuronal glucose transporter that is related to AD[55].

NEUROD6 (neuronal differentiation 6) has the third top regulon in terms of dGRN score (Figure 5F), targeting 24 genes (Figure 5H). It is involved in nervous system development and differentiation[56]. Its downregulation is a biomarker of AD[57] and a significant predictor of cognitive decline[58]. SNPs in its locus are associated with AD in a sex-specific manner[59,60]. ELK1, a member of the TCF subfamily of ETS-domain transcription factors, is another TF whose regulon is in the top five in terms of dGRN score (Figure 5F), comprising 14 genes (Figure 5I). ELK1 is implicated in neuronal differentiation[61] and has been associated with neuronal death and Alzheimer's disease[62]. Elk1 inhibits presenilin 1 (PS1), which is important for making variants of beta amyloid, a trigger for Alzheimer's Disease[63]. Signaling involving Elk1, Ras1 and CentA1 has been reported to connect beta amyloids and synaptic dysfunction, a hallmark of AD[64].

It is also interesting to examine TFs whose predicted regulons have a high dGRN score but a relatively low PsychGRN score, possibly pointing to regulators for which mechanistic evidence is under-represented in PsychENCODE. One such TF is GATA3 (Figure 5F), a pioneer TF[65] involved in signaling pathways associated with neuronal development[66] and control of immune T-cell fate[65]. Donepezil, a drug used for AD, modulates immune response in part by inducing GATA3[67]. GWAS SNPs associated with late onset AD show allele-specific binding of GATA3[68]. Moreover, a GWAS for resilience to cognitive consequences of AD revealed a female-specific locus that interacts with GATA3 and suggested GATA3 as a candidate gene[69].

In summary, differential regulation detected by CIMLA, combined with evidence from the PsychENCODE project, points to regulatory programs that have been reported to play a role in AD and also reveals novel potential regulatory pathologies of AD.

## DISCUSSION

Inference of the differential associations in comparisons of two or multiple groups is of paramount importance in systems biology, with potential applications ranging from tissue-, sex- or population-specific genetic association analysis[70–72] to contrasting the regulatory programs of different populations [26,44]. In recent years, association analysis in genetics and genomics has increasingly relied on complex statistical and ML models [5–9]. However, inference of causal (and differential causal) associations from observational data is fundamentally hindered by confounding variables that are introduced by limitations and biases in data collection or are intrinsic to the problem at hand. This motivated us to approach the inference of differential associations from a causal perspective and develop CIMLA, employing non-linear, multivariable models and model interpretation based on SHAP values[10,23] to approximate a causal estimand of association and changes thereof.

We demonstrated the application of CIMLA for differential gene regulatory network (dGRN) inference from gene expression data. On realistic synthetic data sets CIMLA outperforms existing methods that are based on linear as well as non-linear models, especially in simulations including strong confounders, which are a realistic reflection of biological systems. Our results suggest that CIMLA, in contrast to co-expression methods, can even identify differential regulatory relationships that show relatively small differences in TF-gene correlation between the two populations. Finally, we used CIMLA to infer the differential regulatory program underly Alzheimer's Disease (AD). The resulting dGRN points to CREB3 and NEUROD6 as two important regulators of AD which is in concordance with previous reports for the important role of these TFs in AD [50,57], and also suggests novel potential regulators such as LYL1 and HOXD1 (Figure 5F).

CIMLA employs ML models to impute counterfactuals, i.e., to predict the hypothetical outcome if a sample had belonged to the alternative condition or intervention of interest. This relies on the strong assumption of transferability of the ML models to data distributions that have not been seen during training. While this transferability assumption is consistent with the *positivity* assumption[73], which is crucial for causal inference, the tradeoff between *positivity* and conditional

*ignorability* assumptions[74,75] raises concerns about the transferability of ML models in practice. This defines a future direction for improving CIMLA by employing ML models that can simultaneously learn from the two populations via training jointly on their observational data. (See Shalit et al.[76] and Baur et al.[77].) These advancements in CIMLA may better separate the regulators that are common between conditions from differential associations and also facilitate the extension of this tool to dGRN inference from more than two conditions.

Although we used data from two conditions to train ML models, throughout this study we have used samples of only one of the conditions to quantify local differential scores (Equation (5) in Methods) before aggregating them into a global score (Equation (6) in Methods). However, a rigorous incorporation of samples from both conditions in the local interpretation step can further enhance the performance of dGRN inference and can be a potential direction for improving CIMLA in future studies.

A broader implication of this work lies in the causal interpretation we provide for SHAP values[10,23]. Causal notations were first introduced to SHAP by Janzing et al.[24] and have been later adopted by other studies[10,78]. However, a rigorous mathematical interpretation of the SHAP feature attribution score as the solution to a causal inference problem has been lacking. We bridged this gap by formulating feature attribution as a causal problem and positing a precise set of assumptions under which it simplifies to a statistical quantity that can be approximated from observational data by SHAP. Interestingly, our causal formulation, under three clearly stated assumptions (see Methods), is resolved into a format that is in concordance with the viewpoint of Janzing et al.[24]. But we note that using machine learning and feature attribution methods to draw conclusions about causal relationships needs extreme care. Any of the three stated assumptions can be impacted by problem characteristics and observational data. For example, significant multi-collinearity in data can negatively impact the accuracy of ML models (assumption 1), though this impact can be relieved by regularization techniques during the training of ML models. Data characteristics, including multi-collinearity, might also adversely impact the distributions assumed over causal structures and reference values (assumptions 2 and 3). For example, although the particular distribution assumed over different causal structures guarantees the "local accuracy" property of SHAP[23], recently Kwon and Zou[79] showed that it can lead to suboptimal feature attributions. We hope that the causal interpretation we provided for SHAP and its three underlying assumptions help future studies to further guide feature attribution models toward capturing genuine causal effects.

## METHODS

### A causal estimand for detection of differential regulatory relationships

We present a formulation of GRN reconstruction in the language of causal inference, with the ulterior goal of defining a differential regulatory edge as a causal influence that changes between two groups of samples. For simplicity, the GRN will be reconstructed one target gene at a time, i.e., we will attempt to identify the TFs regulating a given gene; repeating the process for each gene furnishes the full complement of TF-gene regulatory edges in the GRN.

Suppose there are $m$ candidate TFs denoted by indices $M = \{1..m\}$, and the variables $\{X_t\}_{t \in M}$ represent the expression levels of these TFs. Let $Y$ represent the expression of the target gene. Informally, the causal inference problem is to quantify the causal influence of each covariate $X_t$ on $Y$. The common approach to this is to employ the "average treatment effect (ATE)":

$$ATE_t = E[Y|do(X_t = 1)] - E[Y|do(X_t = 0)] \tag{1}$$

where $do(\cdot)$ represents Pearl's "*do*-operator"[80]. In the common case, $do(X_t = 1)$ denotes a binary "treatment" (intervention) such as administering a drug to an individual, while $do(X_t = 0)$ denotes a "control" condition such as administering a placebo. The two expectations are taken over the same population of individuals. This formula also bears resemblance to how a TF's influence on a gene is assessed by a biologist: e.g., the expression of the gene upon knockout of the TF is compared to the wild-type expression, which may loosely be considered as the "$do(X_t = 0)$" and "$do(X_t = 1)$" conditions, respectively. The average effect over biological "samples" (e.g., different cells or different biospecimens) is interpreted as the strength of the TF's regulatory (causal) influence on the gene. We note the similarity of this approach to that of Xing and van der Laan[81].

The above formulation presents an immediate challenge in our context: knocking out a TF $t$ will typically lead to other TFs' levels being affected as well, and an effect may be seen on the target gene due to one of these other TFs regulating the gene, i.e., an indirect effect. However, a TF-gene edge in a GRN is expected to represent direct causal influence, as implemented via the TF binding to enhancers associated with the gene, and indirect effects should not be included. The use of TF binding sites as evidence of direct regulation is the ideal solution to this problem, but not considered here since our goal is to reconstruct GRNs from expression data alone, without access to cis-regulatory information. Thus, we reformulate the above formal definition of the TF's influence on a gene as follows. We first define a "local" version of the ATE that quantifies the effect of a covariate $X_t$ on $Y$ for an arbitrary biological sample $x = \{x_j\}_{j \in M}$. (At this point, $x$ does not necessarily refer to a sample in any given observational data set.) This "Local Treatment Effect" or "LTE" is given by:

$$\text{LTE}_t(x) = E[Y|do(X_t = 1), do(X_{j \in M \setminus \{t\}} = x_{j \in M \setminus \{t\}})] - E[Y|do(X_t = 0), do(X_{j \in M \setminus \{t\}} = x_{j \in M \setminus \{t\}})] \quad (2)$$

where the intervention "$do(X_t)$" is now accompanied by additional interventions "$do(X_{j \in M \setminus \{t\}})$". In other words, to estimate the causal effect of TF $t$ on the target gene in a particular biological sample characterized by TF levels $X = x$, the above definition demands that all TFs other than $t$ be fixed at their levels seen prior to the intervention on $X_t$. We note that such a precise intervention is not likely to be straight-forward or even feasible in practice, but the re-definition in Equation (2) will help clarify the causal inference problem mathematically.

Another challenge with the initial formulation of Equation (1) is the use of a binary "treatment" variable to represent the regulatory action of a TF, whose effect on a gene depends, possibly non-linearly, on its concentration in the cellular context. To address this, we redefine LTE (Equation 2) to be parameterized by $\hat{x}_t$, a baseline level of TF $t$, as follows:

$$\text{LTE}_t(x, \hat{x}_t) = E[Y|do(X_t = x_t), do(X_{j \in M \setminus \{t\}} = x_{j \in M \setminus \{t\}})] - E[Y|do(X_t = \hat{x}_t), do(X_{j \in M \setminus \{t\}} = x_{j \in M \setminus \{t\}})]$$
$$(3)$$

where the effect on the gene in a sample is quantified by the difference between setting $X_t$ to the level observed in the sample ($x_t$) versus setting $X_t$ to the baseline level ($\hat{x}_t$). Note that the baseline level may be greater than or less than $x_t$, corresponding to experimental knock-down or overexpression, respectively, of a TF. Since a single reference value $\hat{x}_t$ may be hard to specify or justify *a priori*, we marginalize it out over a suitable probability distribution $P(\hat{X}_t)$ that may, for instance, be learnt from data. The final definition of the local treatment effect thus becomes:

$$LTE_t(x) = E_{\hat{x}_t \sim P(\hat{X}_t)}[LTE_t(x, \hat{x}_t)] \quad (4)$$

We show in the next section how $LTE_t(x)$ can be estimated from a given observational data set. The procedure will allow LTE estimation for an arbitrary $x$, not only those present in the data set.

Our goal is to define and calculate the *change* of a TF's causal effect on a gene between two conditions (e.g., case and control), represented by a binary variable $C$. To this end, we first define the change in LTE, i.e. "differential LTE", for a single biological sample $x$ as follows:

$$\Delta_t(x) = LTE_t(x|C = 1) - LTE_t(x|C = 0) \tag{5}$$

where $LTE_t(x|C = c)$ is the estimated LTE in condition $c$. Finally, we define the change in a TF's causal influence on a gene for a given population of biological samples $x$ by aggregating over differential LTE for samples in that population:

$$\Lambda_t = \sqrt{E[\Delta_t(x)^2]} \tag{6}$$

i.e., root mean square of changes ($\Delta_t(x)$) in all samples of the population. This is the causal estimand we define as the measure of change in a TF's regulatory effect on a gene, between two conditions.

### A procedure for estimating $\Lambda_t$ from observational data

We assume we are given observational data on the target gene expression $Y$ and the TF expression levels $\{X_i\}_{i \in M}$ in a collection of biological samples (e.g., individual cells or biospecimens), for conditions $C = 1$ and $C = 0$. Let $D_c$ denote the data for condition $C = c$. We present here a procedure that calculates $LTE_t(x|C = c)$ for a given $t \in M$, based on the observational data $D_c$.

First, we use data set $D_c$ to train a machine learning model $f_c$ capable of predicting $Y$ from $X = \{X_i\}_{i \in M}$, i.e., $f_c(X = x) = E_c[Y|X = x]$, where $E_c[\cdot]$ denotes expectation over the distribution underlying data set $D_c$. We then calculate the SHAP value[10,23] $\phi_t(f_c, x)$ that quantifies the contribution of covariate $X_t$ to the model's output at $X = x$. We show (see below) that under certain assumptions, $\phi_t(f_c, x)$ provides an estimate of $E_{\hat{x}_t \sim P_c(\hat{x}_t)}[LTE_t(x, \hat{x}_t)]$, i.e., the SHAP score estimates $LTE_t(x|C = c)$ from the data set $D_c$.

We use the above procedure to estimate $LTE_t(x|C = c)$ for $c = 0$ and $c = 1$ separately (using separate data sets and respective machine learning models $f_0$ and $f_1$), and thus obtain $\Delta_t(x)$ of

Equation (5). We finally estimate $\Lambda_t$ as the root-mean-squared value of $\Delta_t(x)$ over samples $x$ of a data set. Throughout this study, we use only the samples $x \in D_{c=1}$, to estimate the expectation of squared differential LTEs in Equation (6), but future studies might consider using data from both groups. Although it is common to aggregate local SHAP values using a "mean-absolute" function[10], we argue that "root-mean-square" aggregation is more useful in our context (see below).

## SHAP score estimates a causal quantity

As above, assume there exists an observational data set of $m + 1$ variables consisting of a set of covariates (i.e., features) $X_M = \{X_i;\ i \in M = \{1..m\}\}$ and an outcome variable $Y$. We also assume that there exists an unobserved confounder variable $Z'$, that is causally associated with the covariate set $X_M$, but that there are no other unobserved confounders that are associated with the observed covariates $X_M$ and outcome $Y$ (Supplementary Figure S4). Moreover, we allow for causal associations between covariates, with unknown directionality. For each $i \in M = \{1..m\}$, we will assume there exists a causal association between $X_i$ and $Y$ and attempt to quantify it (Supplementary Figure S4, solid arrow), while the associations between $X_{M \setminus \{i\}}$ and $Y$ are not assumed known (Supplementary Figure S4, dashed arrows). With these assumptions, we consider the estimand:

$$\alpha_i(x, \hat{x}_i) \equiv LTE_i(x, \hat{x}_i) = E[Y|do(X_i = x_i, X_{M \setminus \{i\}} = x_{M \setminus \{i\}})] - E[Y|do(X_i = \hat{x}_i, X_{M \setminus \{i\}} = x_{M \setminus \{i\}})] \quad (7)$$

where $\hat{x}_i$ is an arbitrary baseline value of $X_i$. Estimation of $\alpha_i(x, \hat{x}_i)$ is challenging as the underlying causal diagram is not fully resolved. To address this, we borrow ideas from causal discovery[82,83], enumerating all possible causal structures that are consistent with the underlying causal diagram (Supplementary Figure S4), computing $\alpha_i(x, \hat{x}_i)$ for each structure $\psi$ and taking an average. Thus, the estimand of Equation 7 is modified to the following:

$$\alpha_i(x, \hat{x}_i) \equiv LTE_i(x, \hat{x}_i) \equiv E_\psi[LTE_i(x, \hat{x}_i, \psi)] \quad (8)$$

where the expectation $E_\psi[\cdot]$ is taken over a suitable distribution over structures $\psi$ and the notation $LTE_i(x, \hat{x}_i, \psi)$ is introduced to refer to LTE as defined in Equation (3) but under a particular causal structure $\psi$:

$$LTE_i(x, \hat{x}_i, \psi) = E[Y|do(X_i = x_i, X_{M\setminus\{i\}} = x_{M\setminus\{i\}}), \psi] - E[Y|do(X_i = \hat{x}_i, X_{M\setminus\{i\}} = x_{M\setminus\{i\}}), \psi] \qquad (9)$$

There are $2^{m-1}$ distinct causal structures, in all of which feature $X_i$ is causally associated with the outcome $Y$ and in each of which a subset of features $S \subseteq M\setminus\{i\}$ ($0 \leq |S| \leq m-1$) is causally associated with $Y$. Let $A(\psi)$ denotes the subset $S$ of features (excluding feature $i$) associated with $Y$ in causal structure $\psi$ and $NA(\psi)$ denotes the subset $M\setminus(\{i\} \cup S)$ of features that are not associated with $Y$. To simplify Equation (9), we use two lemmas proved in Supplementary Notes S1-3. Using Lemma 1 we can simplify Equation (9) as follows:

$$LTE_i(x, \hat{x}_i, \psi) = E[Y|do(X_i = x_i, X_{A(\psi)} = x_{A(\psi)}), \psi] - E[Y|do(X_i = \hat{x}_i, X_{A(\psi)} = x_{A(\psi)}), \psi]$$

Using Lemma 2 we can further simplify the causal estimand in Equation (9) into a statistical quantity:

$$LTE_i(x, \hat{x}_i, \psi) = E[Y|X_i = x_i, X_{A(\psi)} = x_{A(\psi)}] - E[Y|X_i = \hat{x}_i, X_{A(\psi)} = x_{A(\psi)}]$$

Using law of total expectation, the above equation can be written as:

$$LTE_i(x, \hat{x}_i, \psi) = E_{X_{NA(\psi)}}\left[E[Y|X_i = x_i, X_{A(\psi)} = x_{A(\psi)}, X_{NA(\psi)}]\right]$$
$$- E_{X_{NA(\psi)}}\left[E[Y|X_i = \hat{x}_i, X_{A(\psi)} = x_{A(\psi)}, X_{NA(\psi)}]\right]$$

Given an ML model $f(x)$ that was trained on the data to estimate $f(x) = E[Y|X = x]$, we can further simplify this equation as follows:

$$LTE_i(x, \hat{x}_i, \psi) = E_{X_{NA(\psi)}}[f(X_i = x_i, X_{A(\psi)} = x_{A(\psi)}, X_{NA(\psi)})] - E_{X_{NA(\psi)}}[f(X_i = \hat{x}_i, X_{A(\psi)} = x_{A(\psi)}, X_{NA(\psi)})] \qquad (10)$$

Separately, as in Equation (4), we marginalize the reference value $\hat{x}_i$ in $\alpha_i(x, \hat{x}_i)$ defined in Equation (8), over a distribution $P(\hat{X}_i)$, to define the estimand

$$\alpha_i(x) \equiv E_{\hat{x}_i \sim P(\hat{X}_i)} \alpha_i(x, \hat{x}_i) = E_{\hat{x}_i \sim P(\hat{X}_i)} E_\psi \, LTE_i(x, \hat{x}_i, \psi) = E_\psi E_{\hat{x}_i \sim P(\hat{X}_i)} LTE_i(x, \hat{x}_i, \psi) \qquad (11)$$

where the last equality is obtained by reordering the expectations. If we further assume that a suitable distribution of reference value $\hat{x}_i$ depends on the causal structure $\psi$, and in particular, that $P(\hat{X}_i) = P(X_i|X_{NA(\psi)})$, equation (11) simplifies to:

$$\alpha_i(x) = E_\psi E_{\hat{x}_i \sim P(X_i|X_{NA(\psi)})} LTE_i(x, \hat{x}_i, \psi)$$

$$= E_\psi \left[ E_{\hat{x}_i \sim P(X_i|X_{NA(\psi)})} \left[ E_{X_{NA(\psi)}} [f(X_i = x_i, X_{A(\psi)} = x_{A(\psi)}, X_{NA(\psi)})] \right. \right.$$

$$\left. \left. - E_{X_{NA(\psi)}} [f(X_i = \hat{x}_i, X_{A(\psi)} = x_{A(\psi)}, X_{NA(\psi)})] \right] \right]$$

$$= E_\psi \left[ E_{X_{NA(\psi)}} [f(X_i = x_i, X_{A(\psi)} = x_{A(\psi)}, X_{NA(\psi)})] \right.$$

$$\left. - E_{X_i|X_{NA\{\psi\}}} E_{X_{NA(\psi)}} [f(X_i, X_{A(\psi)} = x_{A(\psi)}, X_{NA(\psi)})] \right]$$

By combining the latter two expectations $E_{X_i|X_{NA\{\psi\}}} E_{X_{NA(\psi)}}$ into an expectation under the joint probability we obtain:

$$\alpha_i(x) = E_\psi \left[ E_{X_{NA(\psi)}} [f(X_i = x_i, X_{A(\psi)} = x_{A(\psi)}, X_{NA(\psi)})] - E_{X_{\{i\} \cup NA(\psi)}} [f(X_{\{i\} \cup NA(\psi)}, X_{A(\psi)} = x_{A(\psi)})] \right] \quad (12)$$

An obvious choice for the probability distribution over $\psi$ is the uniform distribution, i.e., $P(\psi) = \frac{1}{2^{m-1}}$. However, we use $P(\psi) = \frac{1}{m\binom{m-1}{|A(\psi)|}}$, i.e., all causal structures where the same number of covariates $k = |A(\psi)|$ in $X_{M\setminus\{i\}}$ are causally associated with $Y$ together receive a probability of $\frac{1}{m}$ and each such causal structure receives equal probability. With this choice of $P(\psi)$, we obtain:

$$\alpha_i(x) = \sum_{A(\psi) \subseteq M \setminus \{i\}} \frac{1}{m\binom{m-1}{|A(\psi)|}} \left( E_{X_{NA(\psi)}} [f(X_i = x_i, X_{A(\psi)} = x_{A(\psi)}, X_{NA(\psi)})] \right.$$

$$\left. - E_{X_{\{i\} \cup NA(\psi)}} [f(X_{\{i\} \cup NA(\psi)}, X_{A(\psi)} = x_{A(\psi)})] \right)$$

which is equivalent to the local SHAP value of the $i^{th}$ feature in model $f(X)$ at $X = x$ under the adjustment proposed by Janzing et al.[10,24] (see Supplementary Notes S4,5), i.e., $\alpha_i(x) = \phi_i(f, x)$. Thus,

$$\phi_i(f, x) = \alpha_i(x) = E_{\hat{x}_i \sim P(\hat{X}_i)} [\alpha_i(x, \hat{x}_i)] = E_{\hat{x}_i \sim P(\hat{X}_i)} [LTE_i(x, \hat{x}_i)]$$

assuming the modified definition of $LTE_i(x, \hat{x}_i) \equiv E_\psi[LTE_i(x, \hat{x}_i, \psi)]$ (Equation 8) as an average over causal structures $\psi$. Therefore, SHAP[10] provides an estimate for LTE defined in equation (4) under the following assumptions:

1) A machine learning model, $f$, can reliably predict $f(x) = E[Y|X = x]$.
2) Averaging over the $2^{m-1}$ defined causal structures $\psi$ with $P(\psi) = \frac{1}{m\binom{m-1}{|A(\psi)|}}$ reliably estimates LTE, i.e., $LTE_i(x, \hat{x}_i) = E_\psi[LTE_i(x, \hat{x}_i, \psi)]$.
3) $P(X_i|X_{NA(\psi)})$ is a reasonable distribution for the reference value $\hat{x}_i$.

Additionally, *ignorability* (or *unconfoundedness*, i.e., there exist no important unmeasured confounders) and *positivity* (i.e., under any settings of $X_i$, the distribution of the remaining covariates has a similar support), which are the two fundamental assumptions of causal inference[73], are implicitly assumed.

**Aggregating local SHAP values into a global score**

The average of absolute local SHAP values, i.e., $E_x[|\phi_i(x, f)|]$, is commonly used to obtain global SHAP scores[10]. However, here we adopted a root-mean-square aggregation, i.e., $\sqrt{E_x[\phi_i(x, f)^2]}$ -- a second moment of SHAP values. We motivate this choice through a discussion of linear models: when $f$ is a linear model, under the feature independence assumption, we have[23]

$$\phi_i(x, f) = w_i(x_i - E[X_i])$$

where $w_i$ is the coefficient of the $i^{th}$ covariate in the trained model. The root-mean-square of these local SHAP values, assuming all the covariates are normalized to have a unit variance, is:

$$\sqrt{E_x[\phi_i(x, f)^2]} = \sqrt{w_i^2 var(X_i)} = |w_i|$$

which is commonly used as the global feature importance in linear models (assuming normalized features). By extension from this specific case, we aggregate local CIMLA scores $\Delta_t(x)$ (Equation (5)) into a global measure using root-mean-square. We train ML models using normalized features with unit variance and employ equation (6) to obtain: $\Lambda_t = \sqrt{E[\Delta_t(x)^2]}$.

**Synthetic data generation with SERGIO**

We used our previously published single-cell RNA-seq data simulator, SERGIO[36], to generate all synthetic data sets in this study. We first obtained a published GRN from yeast through GeneNetWeaver[37,38], comprising 400 genes of which 20 are master regulators (MR, genes with no upstream regulators) and 17 are non-MR regulators (a total of 37 regulators) and 1155 regulatory interactions. We removed a subset of regulatory interactions in this GRN at random to obtain two sub-GRNs $\mathcal{G}_{C=1}$ and $\mathcal{G}_{C=0}$, corresponding to case and control, sharing $n\%$ of their edges. We repeated this process to obtain 15 GRN pairs with extent of shared edges ranging from $n = 43\%$ to $n = 94\%$. All other parameters necessary for simulation of expression data from these GRNs were selected according to our previously published instructions associated with SERGIO[36]. Each GRN pair was simulated in two settings: low-confounding and high-confounding, as follows. SERGIO simulations require MR profiles as input. An MR profile reflects the expected steady-state expression of MRs in a "cell type". To impose the low-confounding setting, we assumed that simulated cells belong to only one cell type that has the same statistical characteristics (MR profile) in case and control simulations. To achieve this, we generate both case and control profiles, $MR_{C=1}^{low-conf}$ and $MR_{C=0}^{low-conf}$, in one cell-type by sampling all MR levels from the same distribution (specifically, a uniform distribution over the same predefined range for both case and control). On the other hand, to impose a high-confounding setting, we assumed that single cells belong to 10 different cell types that are distinct in case and control simulations. Therefore, we devised two profiles, $MR_{C=1}^{high-conf}$ and $MR_{C=0}^{high-conf}$, in 10 cell-types by sampling MR levels from two distinct distributions in case and control (specifically, uniform distributions over distinct ranges in case and control). Each of the 15 GRN pairs were separately simulated using the MR profiles in the two confounding settings to obtain 15 pairs of single-cell expression profiles in low-confounding and 15 pairs of profiles in high-confounding settings with each profile containing 3000 simulated cells.

To generate noisy data sets, we selected one of the GRN pairs (sharing 94% of their edges) and their corresponding simulated profiles in the high-confounding setting and added various levels (10%, 20%, …, 70%) of dropout using the technical noise module of SERGIO[36]. For each dropout level, we generated five simulated replicates by repeatedly using the SERGIO's noise module.

**Comparing global CIMLA scores of different target genes**
Global CIMLA scores, $\Lambda_t$, for different genes are derived from separate pairs of ML models trained on each gene. This complicates the comparisons between $\Lambda_t$'s of different genes, which are required when prioritizing top differential regulatory edges across all target genes. To facilitate such comparisons across different genes, we standardize target gene expressions to have zero mean and unit variance prior to training ML models. Considering the "local accuracy" property of SHAP values[23], i.e. $f(x) - E[f(X)] = \sum_{i=1}^{m} \phi_i(x, f)$, and the standardization that enforces $E[f(X)] \approx E[Y] = 0$, local SHAP values sum to $f(x)$, which is distributed with a zero mean and unit variance for all genes. This makes the local SHAP values, and hence their difference between case and control groups, as well as their root-mean-square aggregate roughly comparable between different genes. Although this does not provide a mathematical guarantee for comparability of global CIMLA scores of different genes, our benchmarking results on clean simulated data sets suggest reasonable comparability in practice. However, when data are noisy, especially when the noise characteristics are not identical for different genes (for instance, dropout in single-cell RNA-seq has a stronger impact on lowly expressed genes compared to highly expressed ones) the comparability assumption can be weakened. To address this, we also employed a per-gene analysis in our synthetic data evaluations where we assessed the differential regulatory relationships of each gene separately from the others. Similarly, for analysis of the AD data set, we employed a per-gene approach by relying on a background distribution of global CIMLA scores for each gene (see below) to extract the differential regulatory edges of that gene separately from the other target genes.

## Pre-processing of AD snRNA-seq data

The snRNA-seq data we used in this study was obtained from[46] and profiles 48 individuals who were assigned a score between 1-6 representing the final clinical diagnosis of cognitive status at time of death ("cogdx" score). We assigned individuals with a cogdx score of 1 or 2 (no to mild cognitive impairment) to the control group and individuals with a cogdx score of 4 or 5 (Alzheimer's disease as the primary cause of dementia) to the AD group. We excluded the remaining individuals (cogdx score of 3 or 6) from this study since we could not confidently assign them to control or AD groups. Finally, the AD group consists of 30853 single-cells from 22 individuals and the control group consists of 34852 single-cells from 22 individuals. Single-cell expression profiles in each group were separately imputed using MAGIC (with t=2)[45] after a library-size normalization and a "square-root" transformation (as recommended by MAGIC[45]). Genes with low variance in imputed expression (in the bottom 5% in terms of variance over cells of both conditions) were excluded from the target gene sets in the downstream analysis.

## Background distribution of CIMLA scores

In our analysis of AD snRNA-seq data we relied on a background distribution of CIMLA scores reflecting a scenario where we expect no differential regulation. For this, we randomly shuffled the cells between AD and control groups to reconstruct two groups with the same size as the original groups. For each gene and each of the ML model types (RF and NN) we applied CIMLA on this randomly shuffled data set to obtain a background distribution of global CIMLA scores over all TFs. The maximum background score, representing the TF-gene edge with the highest CIMLA score, for a gene and ML type was used as a threshold to filter the differential regulatory edges for that gene and ML type obtained from the original data. For each target gene, a differential regulatory edge whose CIMLA score survives this threshold was assigned a "dGRN score" of $-\log\left(\frac{r}{|TFs|}\right)$ where $r$ denotes the rank of that edge in the sorted list of edges for the gene, and $|TFs|$ denotes the total number of TFs, which is the same for all genes and ML types.

### DEGs in snRNA-seq data

The original analysis of snRNA-seq data[46] published DEGs in each of the six identified cell-types by comparing cells in AD and control. We used the union of the published DEGs in each cell type as a comprehensive DEG set in this study.

### Constructing PsychGRN

We used "GRN1" that was published by the PsychENCODE project[49] and is available from PsychENCODE resource website (http://resource.psychencode.org/). This GRN includes TF-gene edges such that the TF has direct evidence of binding site on the *cis*-regulatory elements of the target gene. TF-gene edges are weighted by the coefficient of the elastic-net regressions trained to predict the target genes expression using the expression of their identified regulators[49]. We constructed PsychGRN by extracting the top 20% of TF-gene edges with the highest absolute edge weights (i.e., regression coefficient). Finally, edge weights in PsychGRN were converted to percentile for better presentation in Figure 5F.

### Details of CIMLA runs

We used two machine learning models in the "ML module" of CIMLA, random forests (RF) and fully connected neural networks (NN). To train RF, CIMLA first performs 3-fold cross-validation (on training data) to select the best hyper-parameter values. These hyper-parameters included the number and maximum depth of decision trees in evaluations on simulated data; and additionally, the "maximum feature" in the case of noisy simulated data. In our analysis of real snRNA-seq data the RF hyper-parameters tuned included the number of decision trees and their maximum depth, minimum samples in leaf nodes and maximum number of leaves. The NN model in CIMLA is a fully connected multi-layer perceptron with 2 (in real data analysis) or 3 (in simulated data analysis) hidden layers and ReLU activation. Additionally, in real data analysis, a dropout layer with $P = 0.5$ was used following the input layer. We used mini-batch training (of size 128) and ADAM optimizer[84] to train the NN models. All ML models were trained in the regression setting using the Mean Square Error loss. For both simulated and real data sets we randomly selected 80% of cells in each group for training ML models, except for training RF on clean and noisy data sets simulated in the high-confounding setting where we used 90% of the data for training. Expressions of the target genes and TFs are normalized to have a zero mean and unit variance prior to training ML models and other downstream analyses. Moreover, when the target gene is also present in the TF list, CIMLA randomly shuffles its corresponding feature column (over cells) to decorrelate the target variable and its corresponding TF column.

CIMLA uses scikit-learn[85] for cross-validation and training RF, and TensorFlow[86] for training NN models. Also, for explaining RF and NN models CIMLA uses TreeSHAP[10] and DeepSHAP[23], respectively. In the case of simulated data sets, we computed and aggregated local CIMLA scores over all training data in one of the groups. For real data analysis we randomly sampled 75% of the training data in AD group for computing and aggregating local CIMLA score.

**Details of dGRN tools compared to CIMLA**

*Co-expression-based methods:* All co-expression methods we used in this study[39–43] were tested using their corresponding implementation in dcanr package[26]. Scores for TF-gene pairs were extracted from the gene-by-gene score matrix output by these implementations. We used the absolute value of TF-gene scores to produce rankings over TF-gene pairs for calculations of performance metrics. The alternative option for producing rankings is to use $-\log(.)$ of adjusted p-values generated by dcanr package which we did not pursue as it results in poorer dGRN inference performances for nearly all co-expression methods and simulated data sets tested (data not shown). EBcoexpress[87] is among the top-performing methods in Bhuva et al. study[26] but was excluded from our analysis as they reported that dGRNs recovered by EBcoexpress and z-score methods are highly comparable.

*BoostDiff:* For every simulated data set, BoostDiff[44] was tested using case and control expression data with a hyper-parameter setting of 100 estimators, 10 features, and 500 subsamples. The true TF list was used as input for regulators in all runs of the tool. Current version of BoostDiff (v0.0.1) outputs two dGRNs corresponding to two different settings of targeting case and control data. For every TF-gene pair, we extracted their maximum score from the two dGRNs and used it to build a ranking over all TF-gene pairs for calculations of performance metrics.

*GENIE3:* This method[31] was used to infer separate GRNs in case and control conditions. For every simulated data set, GENIE3 was separately run on each condition using gene expression data for that condition. For a fair comparison with other methods, the true TF list was used as an input to this method. Weights of TF-gene pairs output by GENIE3 in case and control conditions were used for dGRN inference (GENIE3-diff). A TF-gene pair "*tf,g*" with GENIE3 weights of $w_{C=1}$ and $w_{C=0}$ in case and control conditions receives a differential regulation score as follows:

$$\text{GENIE3-diff}(tf, g) = |w_{C=1} - w_{C=0}|$$

GENIE3-diff scores provide a ranking over TF-gene pairs which reflects their importance in the final dGRN and is used to calculate performance metrics.

## Code and data availability

CIMLA is freely available as a Python package at: https://github.com/PayamDiba/CIMLA. The single-cell transcriptomic data for Alzheimer's disease used in this study was provided by The Rush Alzheimer's Disease Center (RADC) and was obtained from Synapse (https://www.synapse.org/#!Synapse:syn18485175) under the doi 10.7303/syn18485175. All simulated data used in this study (prior to imputation by MAGIC) are available at: https://github.com/PayamDiba/CIMLA_data.

## Acknowledgements

This work was supported by the National Institutes of Health (grant R35GM131819, to Saurabh Sinha) and DOE Center for Advanced Bioenergy and Bioproducts Innovation (U.S. Department of Energy, Office of Science, Office of Biological and Environmental Research under award number DE-SC0018420). People silhouettes used in Figure 1B were obtained from Vecteezy.com.

**Figure 1:** Schematic of CIMLA pipeline. (A) Given $m$ observed covariates ($\{X_1, X_2, ..., X_m\}$) and an outcome of interest ($Y$) in two conditions, case and control, CIMLA's goal is to identify whether the causal association (if any) between a covariate of interest (e.g., $X_i$) and $Y$ has changed between the two conditions. The causal associations between other covariates and $Y$ are generally not known (dashed arrows), and an unobserved confounder ($Z'$) may be involved. (B) Observational data for the outcome of interest ($Y$) and important covariates ($X$) are available for two populations that are different in terms of some condition (e.g., disease status). (C) In each population, the outcome variable is separately modeled as a (non-linear) function of covariates using random forests or neural networks. (D) CIMLA relies on SHAP to approximate a causal measure of association between each covariate and the outcome, in each population. Given a sample from the data and any trained ML model, SHAP searches over possible covariate coalitions to compute the contribution of each covariate to the output of the model locally around the provided sample. This "local SHAP score" is obtained for all provided samples, resulting in matrices of local SHAP scores. (E) The population-specific local SHAP scores of covariates, obtained in the previous step, are compared between the two populations and aggregated over samples into a global CIMLA score for each covariate, representing the strength of differential association between that covariate and the outcome variable. The global CIMLA scores are used to obtain a ranking over covariates by their differential association with outcome.

**Figure 2:** Benchmarking on clean simulated data. (A) Two of the 15 dGRNs used for simulations of gene expression data and benchmarking of GRN reconstruction in this study. The dGRNs shows are those with the least (43%) and most (94%) fraction of shared regulatory edges between conditions. Gray and colored edges represent shared and condition-specific edges, respectively. (B) Underlying causal diagram for the low-confounding simulation setting including master regulators (MR), non-MR TFs and target gene of interest. Node $g$ represents the target gene and nodes $tf_{i \in \{1..m\}}$ represent transcription factors, some of which are master regulators (dark blue nodes). Node $C$ represents disease status, which can impact the associations between a subset of TFs and the target gene (marked with diamonds hit by green arrows). We used MR profiles drawn from the same distribution for simulations of the two conditions, to rule out extraneous associations between $C$ and $g$ through backdoor paths. (C) Underlying causal diagram for the high-confounding simulation setting. We used different MR profiles (in 10 cell-types) for the simulations of the two conditions to mimic an unobserved confounder, $Z'$, that controls both $C$ and MR expressions. Here, in contrast to (B), there exist backdoor paths of association between $C$ and $g$ (which pass through $Z'$), thus creating a confounding effect for detection of direct causal associations between $C$ and TF-target gene associations. (D) Area under the receiver operating characteristic curve (AUROC) for CIMLA and co-expression methods on low-confounding simulated data. (E) Area under the precision-recall curve (AUPRC), normalized by the expected AUPRC of a random classifier, for CIMLA and co-expression methods. In (D) and (E), each box plot shows performance measures for the 15 data sets. (F,G) Performance of CIMLA versus z-score-P on each of the low-confounding data sets, in terms of (F) AUROC, and (G) normalized AUPRC. (H) Performance of CIMLA and z-score method for varying levels of similarity (shared edges) between the case and control GRNs. (I,J) Performance of CIMLA and co-expression methods on high-confounding simulated data sets, in terms of (I) AUROC, and (J) normalized AUPRC. (K,L) Performance of CIMLA versus z-score-S on each of the high-confounding simulated data sets, in terms of (K) AUROC, and (L) normalized AUPRC. (M,N) Comparison of CIMLA with non-linear, multivariable modeling methods on the high-confounding simulated data sets, in terms of (M) AUROC, and (N) normalized AUPRC.

**Figure 3:** (A) Distributions of delta-correlations (dCORR, difference of TF-gene correlation coefficient between case and control populations), for true differential and non-differential edges, across all 15 simulated data sets, in low- and high-confounding settings. The non-differential edges consist of the TF-gene edges that are present in (shared by) both case and control GRNs, as well as those that are absent in both GRNs. (B,C) Relationships between delta-correlation (dCORR) and scores (converted to percentile) assigned to the true differential edges by CIMLA (blue) and z-score-S (burgundy) methods in (B) low-confounding, and (C) high-confounding settings. Panel (C) additionally shows z-score-S assigned scores from the low-confounding setting for comparison. Insets show zoomed-in views of the top 10 percentile scores suggesting that in high-confounding setting, CIMLA assigns high scores to differential pairs with even smaller delta-correlations compared to low-confounding setting, while z-score-S demands relatively larger delta-correlations to assign high scores. (D,E) True Positive Rate (TPR) for the top 5% predictions of CIMLA and z-score-S for the task of discriminating differential edges from non-differential edges. Evaluations are done separately for all TF-gene pairs with small delta-correlations ($dCorr \leq 0.16$) and those with large delta-correlations ($dCorr > 0.16$), shown as two groups in (D) low-confounding, and (E) high-confounding settings. The 0.16 cutoff was used based on the median of delta-correlations over the union of true differential edges in all low- and high-confounding simulated datasets.

**Figure 4:** Benchmarking on noisy simulated data. (A,B) Performance of CIMLA and other methods in terms of (A) AUROC, and (B) Normalized AUPRC, at varying dropout levels. Performance is assessed for each gene separately and each bar represents the median over all differentially regulated genes (every gene with at least one true differential regulation was included) and five simulated replicates. (C) Pearson's correlation between CIMLA-RF and CIMLA-NN scores of all candidate transcription factors of each target gene, at different dropout levels. At each dropout level, distribution of correlation values is shown over all target genes and simulated replicates. (D) Relationship between CIMLA-NN and CIMLA-RF scores across all TF-gene pairs and all (noisy) simulated data sets. Scores in each simulated dataset were max-normalized (with respect to the maximum score in that dataset) for both methods separately. (G) Performance (F1 score) of dGRN inference derived from the intersection or union of the top 10% scoring TF-gene pairs predicted by CIMLA-RF and CIMLA-NN is compared with the performance of the individual methods and z-score-S method at their top 10% scoring TF-gene pairs.

**Figure 5:** Case study of Alzheimer's disease. (A) Training and test data performance (R-squared, $R^2$) of the RF and NN models trained in step 1 of CIMLA for each of the target genes, in the AD and control groups. (Genes with negative test R-squared are not shown. Note that R-squared for nonlinear regression can have negative values indicating significant overfitting.) (B) Comparing the test performance (R-squared, $R^2$) of NN with that of RF for genes that are reliably modeled ($R^2 \geq 0.5$ on both train and test data in both AD and control groups) by both models. NN yields more accurate predictions than RF in general, suggesting that it is less prone to overfitting. (C) Distributions of the number of differential regulatory edges found for each target gene (after thresholding by maximum background score) by CIMLA-NN and CIMLA-RF. (D) Top 20 "hub" TFs (those targeting greatest number of DEGs) found by CIMLA-NN. (E) Top 20 differentially regulated DEGs, i.e., those with greatest number of differential regulators (dREGs) found by CIMLA-NN. (F) Summary of the top 40 largest "regulons" (targets of individual TFs) in the dGRN of AD obtained by intersecting the predictions of CIMLA-union with PsychGRN. For each regulon, its dGRN score (average of dGRN scores of the TF-gene pairs in the regulon), PsychGRN score (average of PsychGRN percentile scores of the TF-gene pairs in the regulon) and its size (number of target genes in the regulon) are shown. (G-I) Predicted regulons for (G) CREB3, (H) NEUROD6, and (I) ELK1 transcription factors. Predicted target genes that are differentially expressed between AD and control groups (i.e. DEGs) are shown by orange-colored nodes. Edges are labeled by their CIMLA dGRN score.

Figure 1

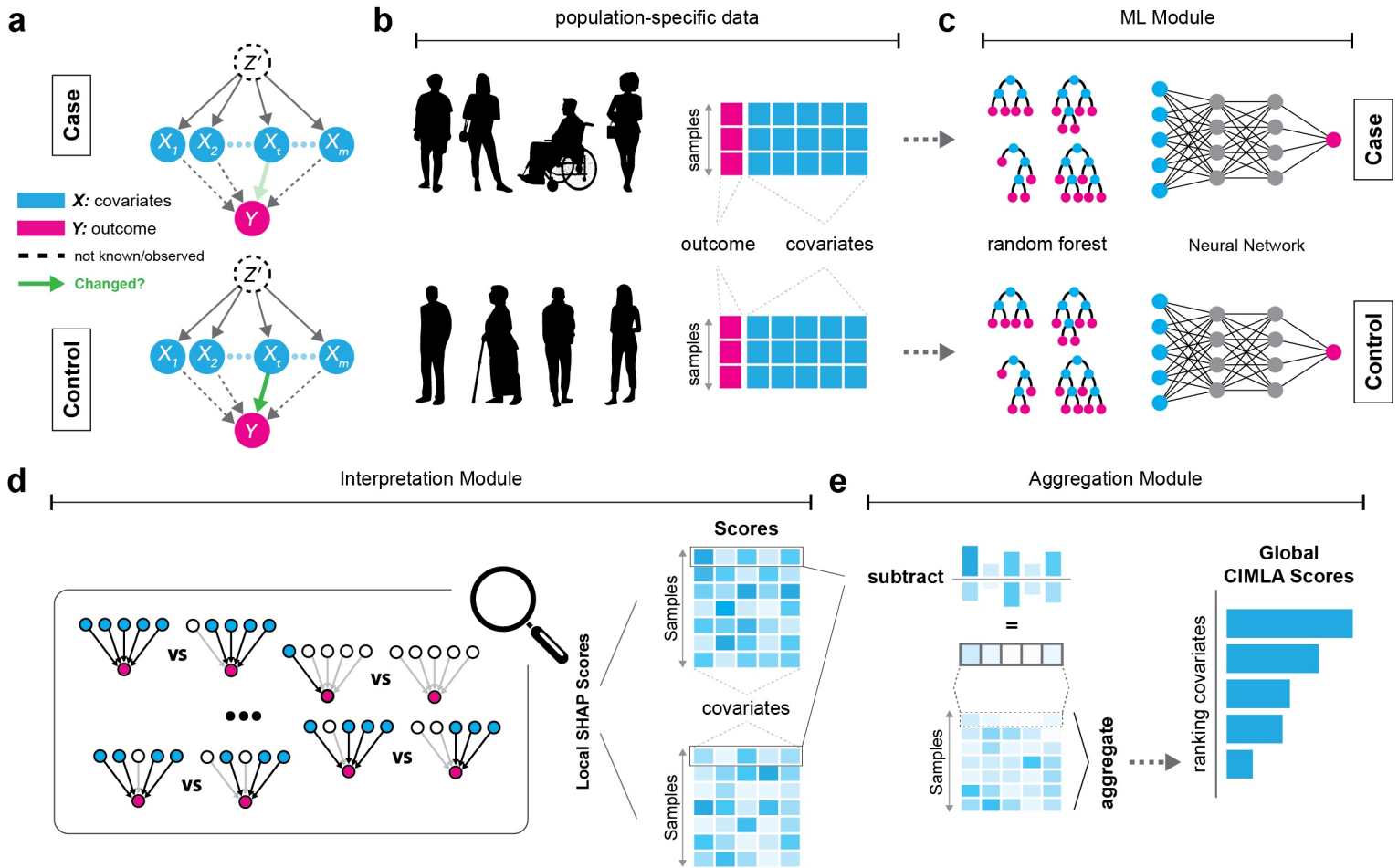

**Figure 2**

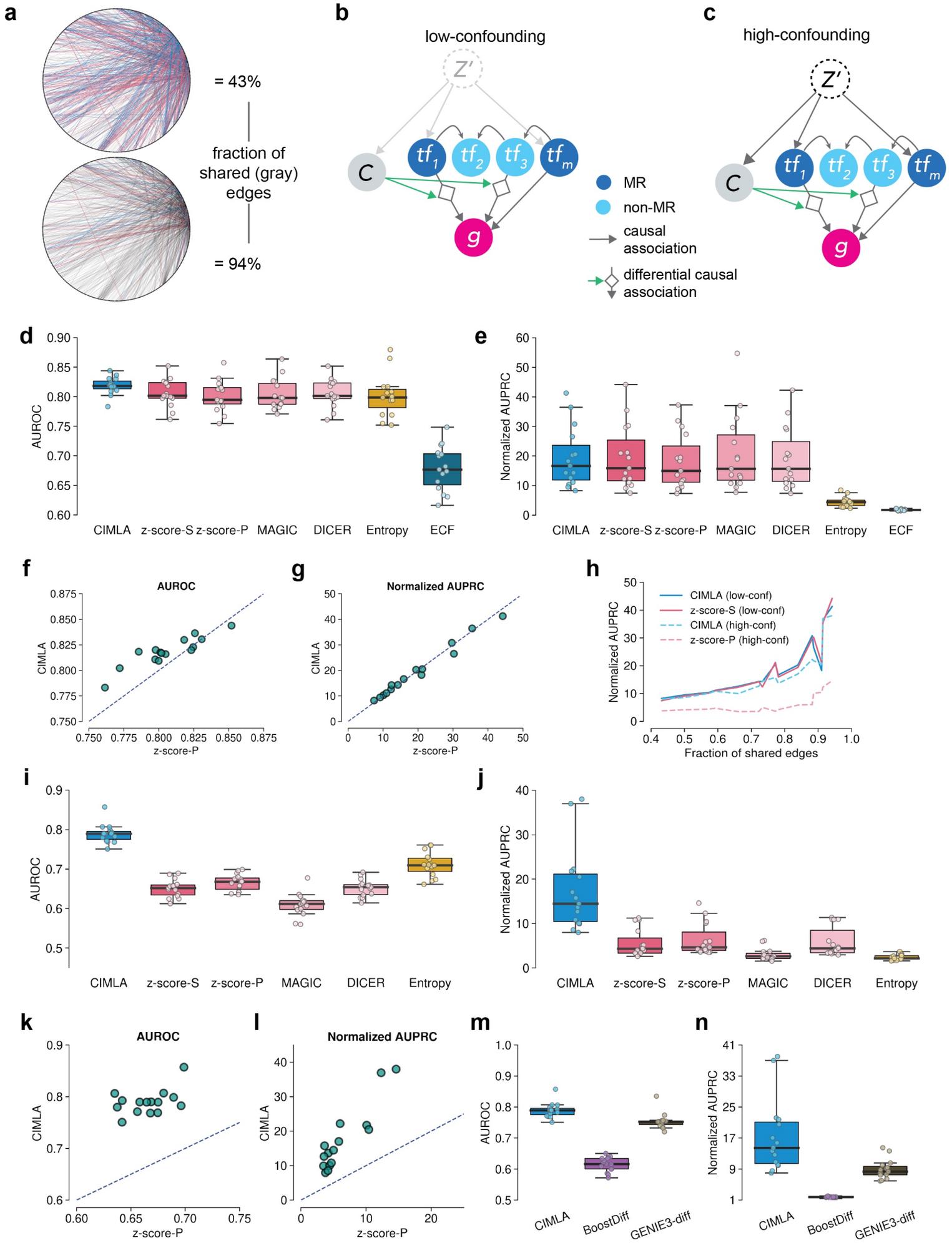

# Figure 3

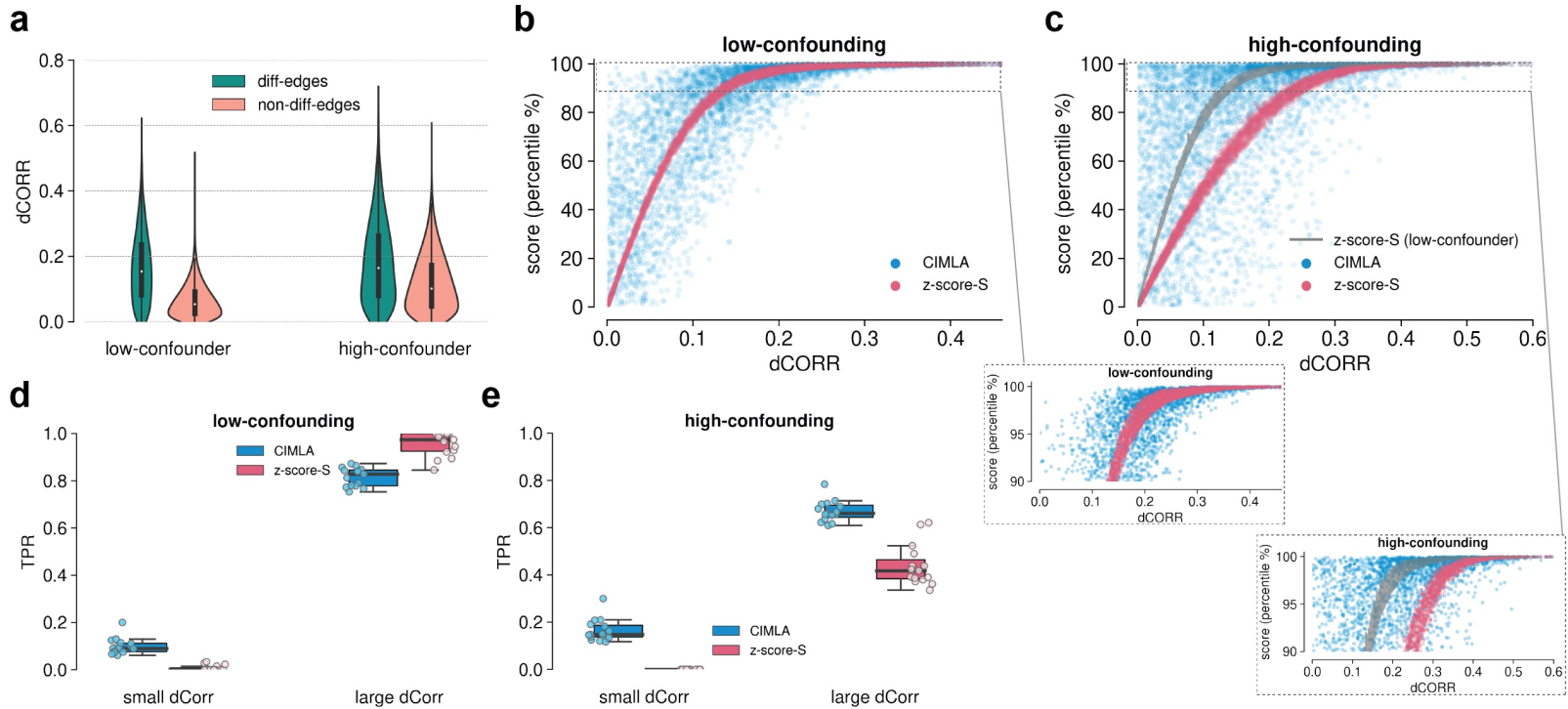

Figure 4

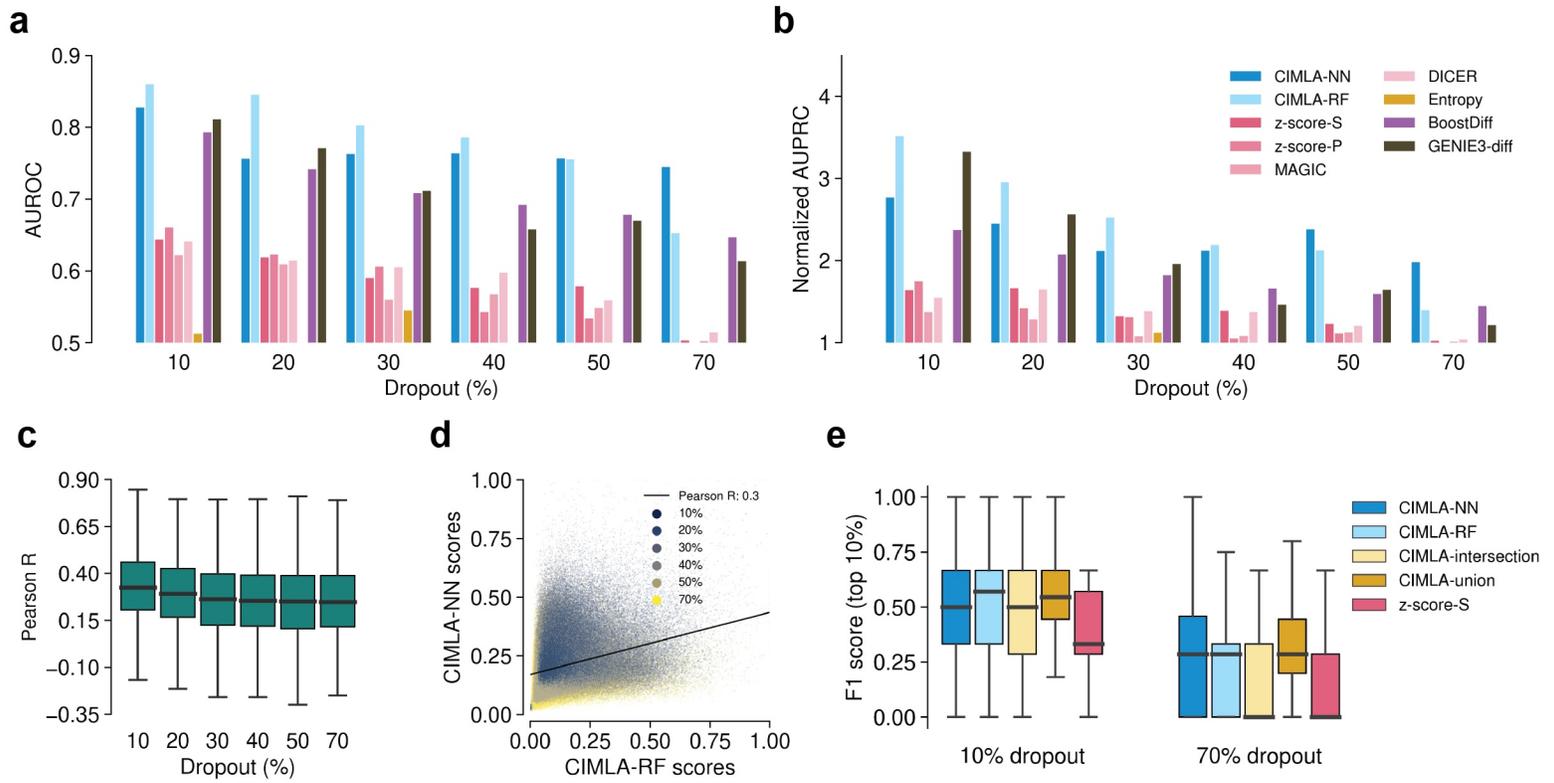

Figure 5

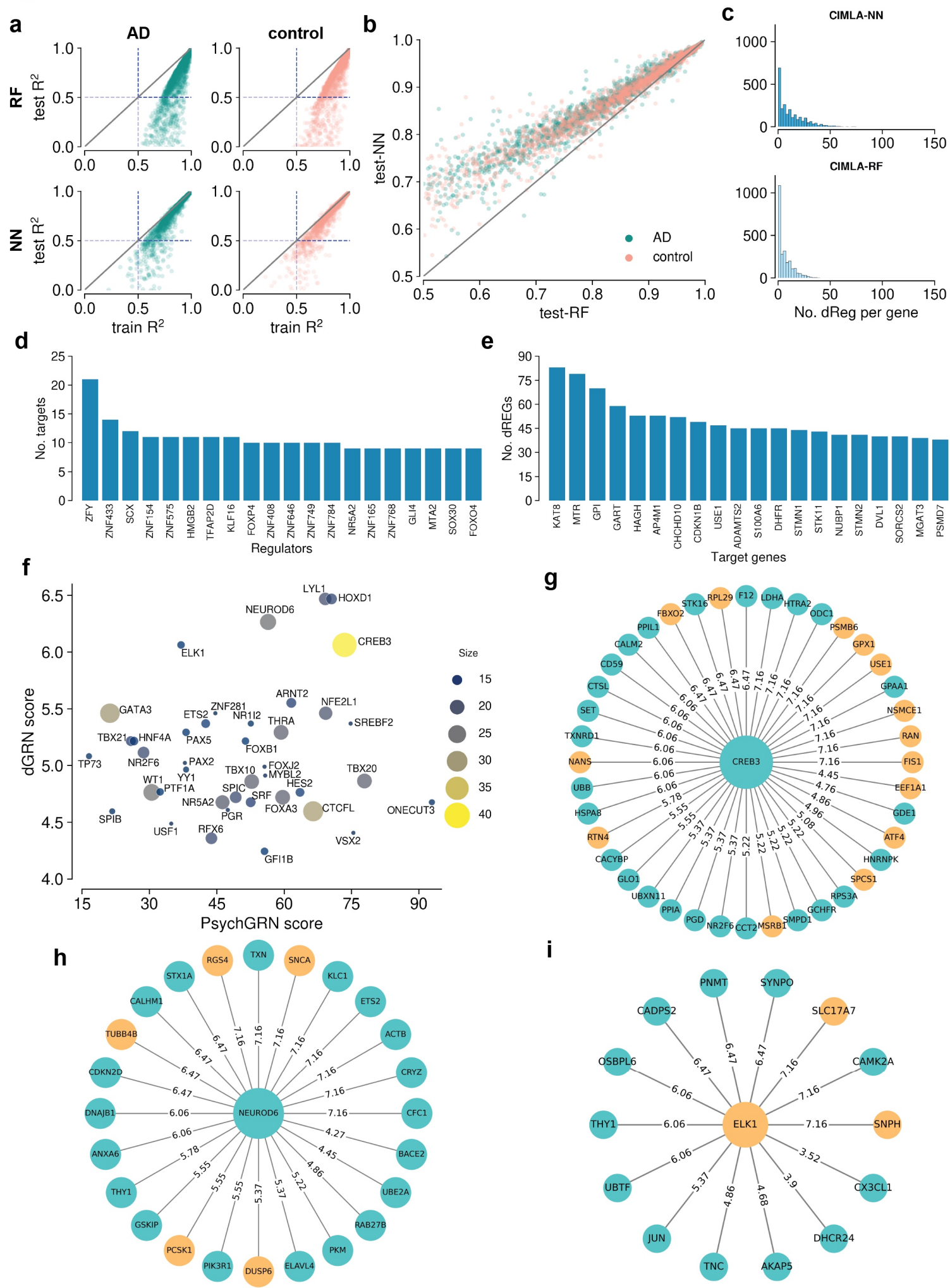

# Supplementary Information

## CIMLA: Interpretable AI for inference of differential causal networks

Payam Dibaeinia[1], Saurabh Sinha[2*]

[1] Department of Computer Science, University of Illinois at Urbana-Champaign, Urbana, IL, 61801, USA
[2] Wallace H. Coulter Department of Biomedical Engineering, Georgia Institute of Technology, Atlanta, GA, 30332, USA

[*] Correspondence: saurabh.sinha@bme.gatech.edu

# Supplementary Figures

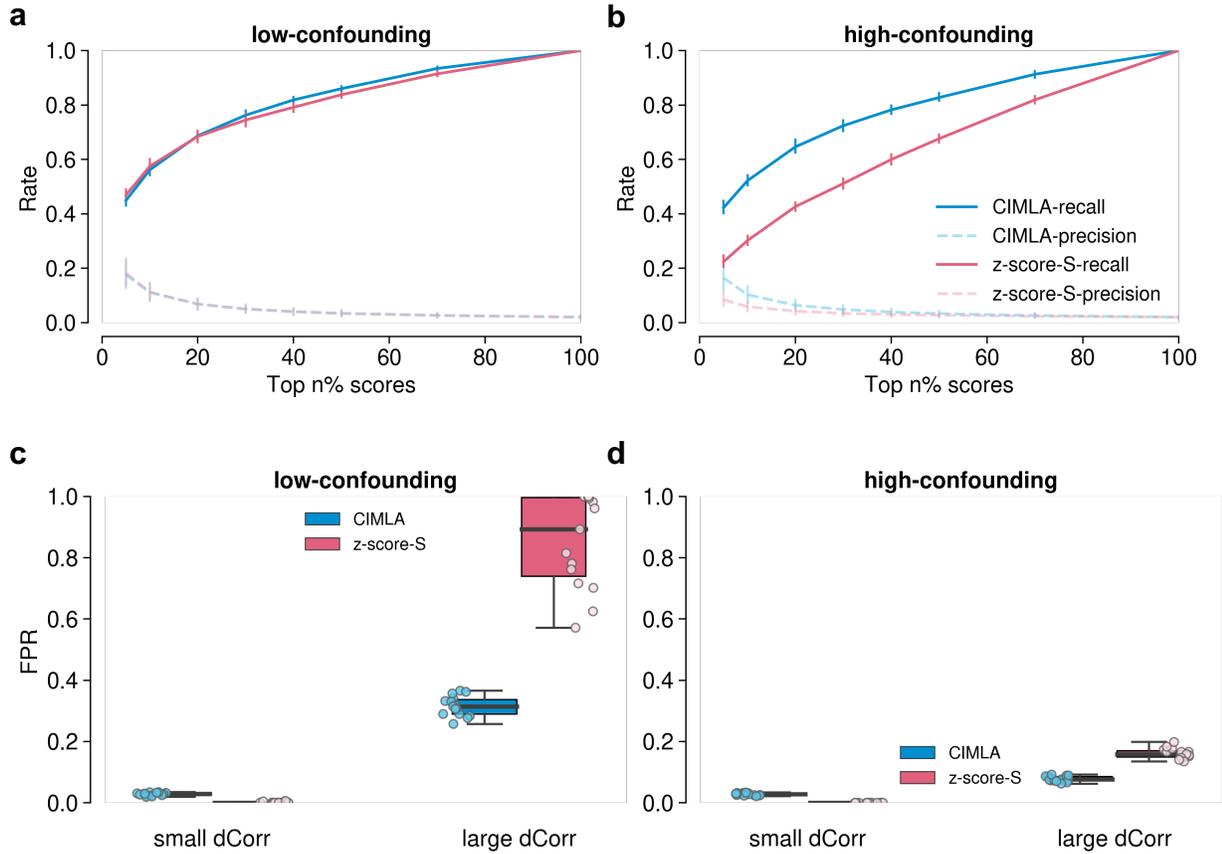

**Figure S1**: (A,B) Precision and recall rates for various cutoffs on the top predictions made by CIMLA and z-score-S on (A) low-confounding, and (B) high-confounding simulated data (error bars reflect variations among 15 simulated datasets). (C,D) False Positive Rate (FPR) for the top 5% predictions of CIMLA and z-score-S for the task of discriminating differential edges from non-differential edges. Evaluations are done separately for all TF-gene pairs with small delta-correlations ($dCorr \leq 0.16$) and those with large delta-correlations ($dCorr > 0.16$), shown as two groups in (C) low-confounding, and (D) high-confounding settings.

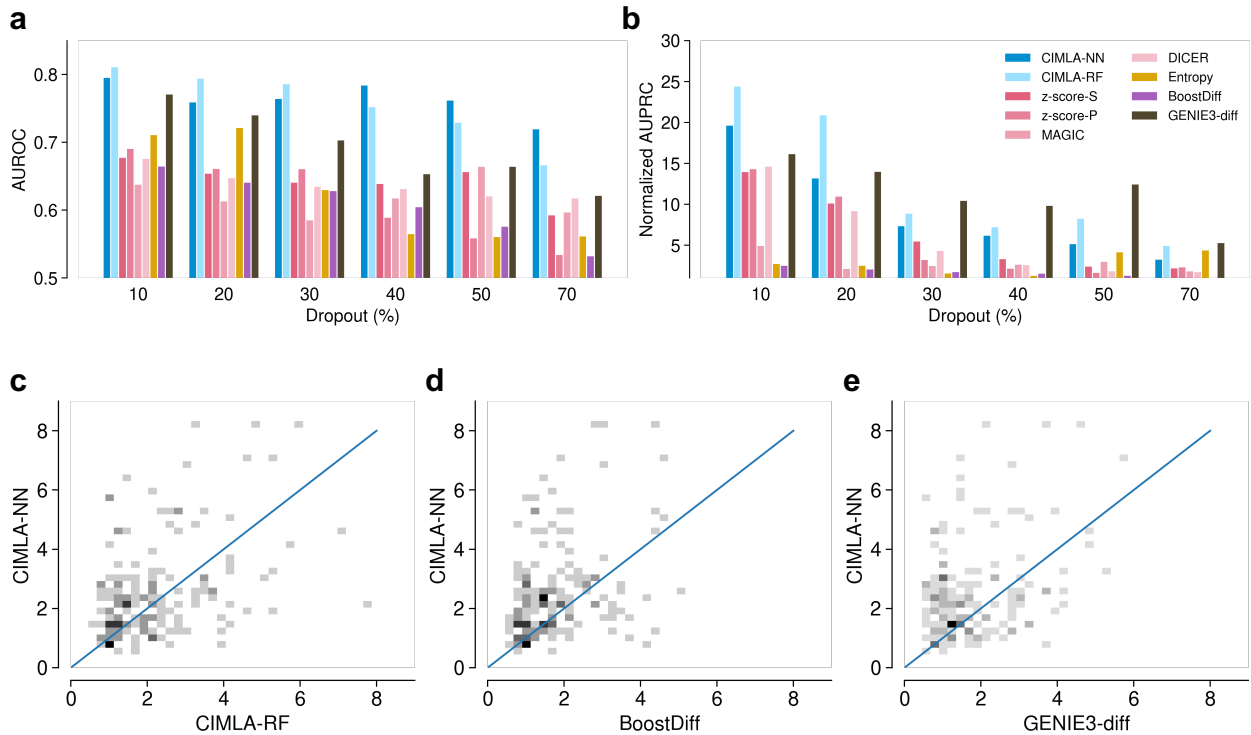

**Figure S2**: (A,B) Performance of CIMLA and other methods for the "entire dGRN" prediction task on noisy simulated data at varying dropout levels in terms of (A) AUROC, and (B) Normalized AUPRC. Each bar represents the median over five simulated replicates. (C-E) Comparing the performance of different methods in terms of normalized AURPC for predicting the differential regulations of each gene in the "per-gene" prediction task. Darker colors show higher counts of genes in the 2-dimensional histogram.

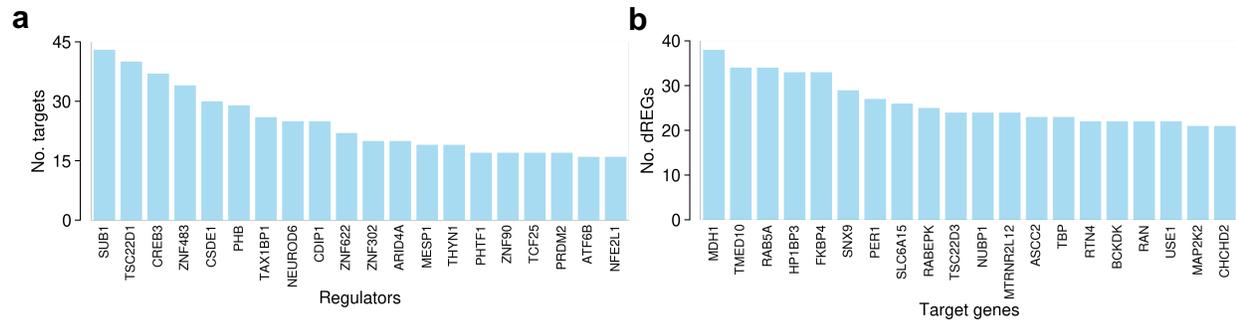

**Figure S3**: (A) Top hub TFs, i.e., those targeting greatest number of DEGs, in the dGRN found by CIMLA-RF. (B) Top differentially regulated DEGs, i.e., those with greatest number of differential regulations (dREGs) found by CIMLA-RF.

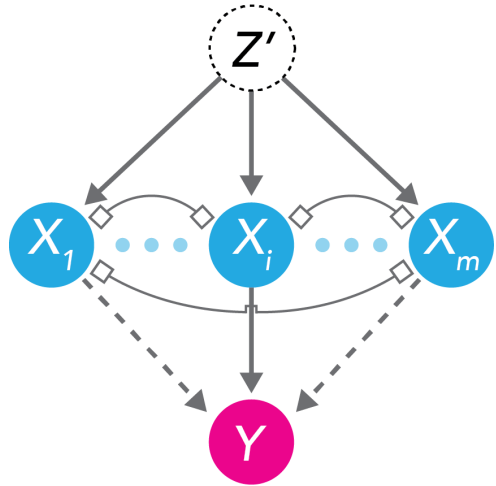

**Figure S4**: A causal diagram for feature attribution problem. The diagram includes $m$ observed covariates, $\{X_i\}_{i \in \{1..m\}}$, and an outcome of interest, $Y$. The goal is to identify the causal association between covariate $X_i$, and $Y$ (solid arrow), while the associations between other covariates and $Y$ are not known (dashed arrows). Causal dependencies among covariates are assumed with unknown directionality (lines with square ends). Confounder variable $Z'$ is not observed and is assumed to be causally associated with all covariates but not with the outcome.

## Supplementary Notes

### Note S1: Rules of do-calculus

Assume all the causal associations between pairs of variables are represented by a Directed Acyclic Graph (DAG) $G$. Let $Y$ denote the outcome variable of interest and $T$, $Z$, and $W$ represent three disjoint set of nodes, excluding $Y$, in $G$. Also, let $G_{\bar{V}}$ represent the graph obtained by removing all the incoming edges of $V$, where $V$ denotes a subset of nodes in $G$, and $G_{\underline{V}}$ represent the graph obtained by removing all the outgoing edges of $V$. Using these notations, we may write the second and third rules of do-calculus[1] as:

*Rule 2:*

$$P(Y|do(T), do(Z), W) = P(Y|do(T), Z, W) \quad \text{if} \quad Y \perp\!\!\!\perp_{G_{\bar{T}\underline{Z}}} Z \mid T, W \quad \text{(S1)}$$

The notation $Y \perp\!\!\!\perp_{G_{\bar{T}\underline{Z}}} Z \mid T, W$ should be read as: *Y and Z are d-separated by conditioning on $T \cup W$ in graph $G_{\bar{T}\underline{Z}}$*, i.e., the graph obtained by removing all the incoming edges of $T$ and all the outgoing edges of $Z$. Two variables $a$ and $b$ are "d-separated" by $c$ if there exists no collider-free path between $a$ and $b$ in the causal graph that does not visit $c$, i.e., if $c$ "blocks" every collider-free path between the variables. (A "collider" is a node in a graph where two arrows "collide head-to-head".)

*Rule 3:*

$$P(Y|do(T), do(Z), W) = P(Y|do(T), W) \quad \text{if} \quad Y \perp\!\!\!\perp_{(G_{\bar{T}})\overline{Z(W)}} Z \mid T, W \quad \text{(S2)}$$

The notation $Y \perp\!\!\!\perp_{(G_{\bar{T}})\overline{Z(W)}} Z \mid T, W$ should be read as: *Y and Z are d-separated by conditioning on $T \cup W$ in graph $(G_{\bar{T}})\overline{Z(W)}$*, i.e., the graph obtained by first removing all incoming edges to $T$, obtaining the graph $G_{\bar{T}}$, and then removing all incoming edges into the set $Z(W)$, defined as the set of nodes in $Z$ that are not an ancestor of any node in $W$ in graph $G_{\bar{T}}$.

Informally speaking, rule 2 of do-calculus allows the conversion of a *do*-term to a conventional statistical conditioning and rule 3 allows the elimination of a *do*-term from the causal quantity.

## Note S2: Lemma 1

*Lemma 1:* Consider the causal structure $\psi$ shown as a DAG $G$ in Figure S5A, where a subset of the variables $\{X_j\}$ influence the outcome variable $Y$, and the variables $\{X_j\}$ are in turn influenced by a variable $Z'$. The variable $X_i$ influences $Y$, $A(\psi)$ represents the indices of other variables that are associated with $Y$, and $NA(\psi)$ represents the indices of variables that are not associated with $Y$. The union of all directed associations from variables in $X_{NA(\psi)}$ to nodes in $X_{A(\psi)\cup\{i\}}$ (resp. from nodes in $X_{A(\psi)\cup\{i\}}$ to nodes in $X_{NA(\psi)}$) is shown by a directed edge from $X_{NA(\psi)}$ to $X_{A(\psi)\cup\{i\}}$ (resp. from $X_{A(\psi)\cup\{i\}}$ to $X_{NA(\psi)}$). (Note: although the directed edges shown between $X_{A(\psi)\cup\{i\}}$ and $X_{NA(\psi)}$ suggest the existence of a cycle, the actual directed edges represented by these two edges are assumed not to form a cycle.) For this causal structure,

$$E[Y|do(X_{A(\psi)\cup\{i\}} = x_{A(\psi)\cup\{i\}}), do(X_{NA(\psi)} = x_{NA(\psi)})] = E[Y|do(X_{A(\psi)\cup\{i\}} = x_{A(\psi)\cup\{i\}})]$$

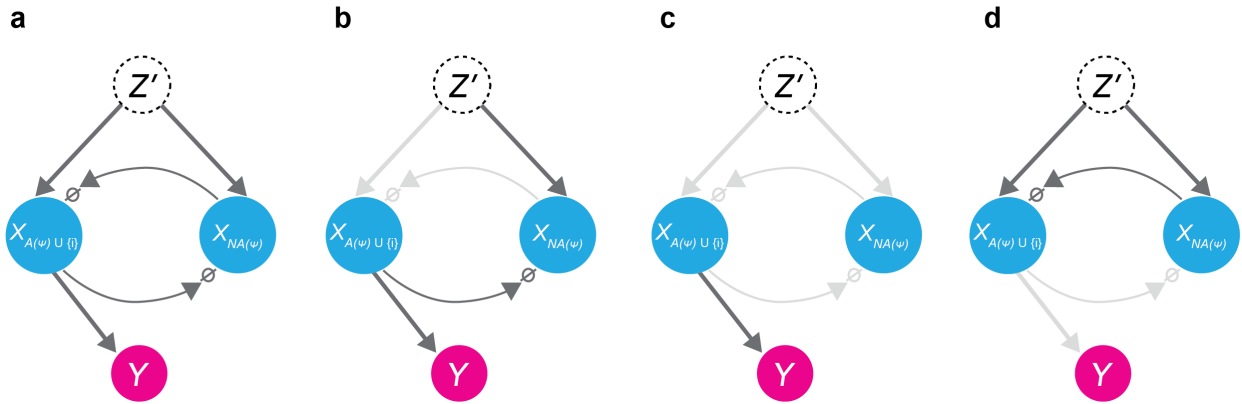

**Figure S5**: (A) Causal DAG, $G$, representing the causal structure $\psi$ consistent with graph in Supplementary Figure S4. Associations between $X_{A(\psi)\cup\{i\}}$ and $X_{NA(\psi)}$ are represented by two "→ ∅" edges to emphasize that they do not introduce cycles. (B) Graph $G_{\overline{T}}$ in the notations used for the proof of Lemma 1. (C) Graph $(G_{\overline{T}})_{\overline{Z(W)}}$ in the notations used for the proof of Lemma 1. (D) Graph $G_{\overline{T}\underline{Z}} = G_{\underline{Z}}$ in the notations used for the proof of Lemma 2.

*Proof:* This lemma is a direct result of the third rule of do-calculus. We define $Z = X_{NA(\psi)}$, $T = X_{A(\psi)\cup\{i\}}$, and $W = \emptyset$; thus, in $G$, all direct edges into $Y$ are from $T = X_{A(\psi)\cup\{i\}}$ and none from $Z = X_{NA(\psi)}$. We first obtain graph $G_{\overline{T}}$ of equation (S2) by removing edges incoming into $T = X_{A(\psi)\cup\{i\}}$ (Figure S5B). Since $W$ is empty, we have $Z(W) = Z$, where $Z(W)$ is defined as the set of nodes in $Z$ that are not an ancestor of any node in $W = \emptyset$. Thus, $(G_{\overline{T}})_{\overline{Z(W)}}$ is obtained from $G_{\overline{T}}$ by further removing all edges coming into $Z = X_{NA(\psi)}$, giving us the graph shown in Figure S5C. Note that this derivation of $(G_{\bar{T}})_{\overline{Z(W)}}$ also removes all causal dependencies among variables $\{X_j\}$, regardless of their direction. Thus, in the resulting graph shown in Figure S5C, there is no path of association between $Z = X_{NA(\psi)}$ and $Y$ and therefore the condition of the third rule of do-calculus is satisfied and we get:

$$E[Y|do(T), do(Z), W] = E[Y|do(T), W]$$

and thus, by substituting the definitions $T = X_{A(\psi) \cup \{i\}}$, $Z = X_{NA(\psi)}$, $W = \emptyset$, we have

$$E[Y|do(X_{A(\psi) \cup \{i\}} = x_{A(\psi) \cup \{i\}}), do(X_{NA(\psi)} = x_{NA(\psi)})] = E[Y|do(X_{A(\psi) \cup \{i\}} = x_{A(\psi) \cup \{i\}})] \quad (S3)$$

This completes the proof of Lemma 1.

**Note S3: Lemma 2**

*Lemma 2*: Consider the causal structure $\psi$, represented by DAG $G$ (Figure S5A), with variables and associations as defined in Lemma 1. Then

$$E[Y|do(X_{A(\psi)\cup\{i\}} = x_{A(\psi)\cup\{i\}})] = E[Y|X_{A(\psi)\cup\{i\}} = x_{A(\psi)\cup\{i\}}]$$

*Proof:* This Lemma is a direct result of the second rule of *do*-calculus. Using the notations of equation (S1), we define $Z = X_{A(\psi)\cup\{i\}}$, and $T = W = \emptyset$. We obtain graph $G_{\overline{T}\underline{Z}} = G_{\underline{Z}}$ of equation (S1) by removing edges outgoing from $Z = X_{A(\psi)\cup\{i\}}$ in $G$. Since in the resulting graph shown in Figure S5D, there is no path of association between $Z = X_{A(\psi)\cup\{i\}}$ and $Y$, the condition of the second rule of *do*-calculus is satisfied and we get:

$$E[Y|do(T), do(Z), W] = E[Y|do(T), Z, W]$$

and thus, by substituting the definitions $T = W = \emptyset$, $Z = X_{A(\psi)\cup\{i\}}$, we have

$$E[Y|do(X_{A(\psi)\cup\{i\}} = x_{A(\psi)\cup\{i\}})] = E[Y|X_{A(\psi)\cup\{i\}} = x_{A(\psi)\cup\{i\}}] \qquad (S4)$$

This completes the proof of Lemma 2.

**Note S4: Shapley Value for explaining a multi-player game**

Assume $M = \{1..m\}$ is a set of $m$ players contributing to a game whose outcome is measured by a value function $v$:

$$v(S) \in \mathbb{R}, \quad \forall S \subseteq M \quad \text{and} \quad v(\emptyset) = 0$$

Contribution of player $i \in M$ to the game's outcome can be computed by Shapley value[2] of the $i^{th}$ player:

$$\phi_i = \Sigma_{S \subseteq M \setminus \{i\}} \frac{1}{m\binom{m-1}{|S|}} \left( v(S \cup \{i\}) - v(S) \right) \tag{S5}$$

Shapley value defined as above is a fair contribution allocation method as it satisfies several desirable properties discussed elsewhere[2,3].

**Note S5: SHAP Value for explaining an arbitrary function**

Next, the question is how to use Shapley value for explaining the output of an arbitrary function. Assume we have an observational dataset $\mathcal{D} = \{(\mathcal{D}_X, \mathcal{D}_Y)_j\}_{j \in \{1..n\}} = \{(\{X_i\}_{i \in M}, \{Y\})_j\}_{j \in \{1..n\}}$, where $M = \{1..m\}$, containing $n$ observations for an outcome of interest $Y$ and $m$ covariates $X = \{X_i\}_{i \in M}$ that are independent and identically distributed (IID) samples from a distribution $P(X, Y)$, $X \in \mathbb{R}^m$, $Y \in \mathbb{R}$. Also, assume that we have trained a machine learning model $f$ on the data through regression to predict $Y$ from $X$, so we have:

$$\forall x \sim P(X), \quad f(x) = E[Y|X = x]$$

where $P(X)$ denotes the marginal distribution of covariates. The goal is to measure the contribution of the $i^{th}$ dimension (henceforth called "feature $i$") of input $x \in \mathcal{D}_X$ to the function's output at $X = x$, i.e., $f(x)$. In order to employ the idea underlying equation (S5), we need to define how a function $f$ may be evaluated on a subset $S \subseteq M$ of features, i.e., $f(x_S)$. By having a well-defined notion for $f(x_S)$ we can extend equation (S5) to measure the contribution of feature $i$ to the output of function $f$ locally at $X = x$ as follows:

$$\phi_i(f, x) = \sum_{S \subseteq M \setminus \{i\}} \frac{1}{m \binom{m-1}{|S|}} \left( f(x_{S \cup \{i\}}) - f(x_S) \right) \quad (S6)$$

Lundberg and Lee[3] proposed to define

$$f(x_S) = E[f(X)|X_S = x_s] = E_{P(X_{\bar{S}}|X_S = x_S)}[f(x_S, X_{\bar{S}})]$$

where $\bar{S} = M \setminus S$. Later, Janzing et al.[4] modified this notion by defining $f(x_S) = E[Y|do(X_S = x_s)]$ which under their assumed causal graph is simplified to $E[Y|do(X_S = x_s)] = E_{P(X_{\bar{S}})}[f(x_S, X_{\bar{S}})]$, where $do(.)$ represents the Pearl's do-operator[5]. This interpretation of SHAP values was later adopted by Lundberg et al.[6]. By using the notion proposed by Janzing et al. in equation (S6) we obtain:

$$\phi_i(f, x) = \sum_{S \subseteq M \setminus \{i\}} \frac{1}{m \binom{m-1}{|S|}} \left( E_{X_{\bar{S} \setminus \{i\}}} f\left(x_{S \cup \{i\}}, X_{\bar{S} \setminus \{i\}}\right) - E_{X_{\bar{S}}} f(x_S, X_{\bar{S}}) \right) \quad (S7)$$

In order to relate to the language used in Methods, we define $S = A(\psi)$, where $\psi$ is the causal structure in which only the subset $S \subseteq M\setminus\{i\}$ of covariates (i.e., $X_S$) and feature $i$ (i.e., $X_i$) are associated with outcome $Y$. Thus, we have $\overline{S} = NA(\psi) \cup \{i\}$. By substituting these notations in equation (S7) we get:

$$\phi_i(f, x) = \sum_{A(\psi) \subseteq M\setminus\{i\}} \frac{1}{m \binom{m-1}{|A(\psi)|}} \left( E_{X_{NA(\psi)}} f(X_i = x_i, X_{A(\psi)} = x_{A(\psi)}, X_{NA(\psi)}) \right.$$

$$\left. - E_{X_{NA(\psi) \cup \{i\}}} f(X_{A(\psi)} = x_{A(\psi)}, X_{NA(\psi) \cup \{i\}}) \right)$$

which is equivalent to our approximation, $\alpha_i(x)$, formalized in Methods, for estimating local causal feature association defined by $LTE_i(x)$ in equation (4) in Methods. Therefore, SHAP value[3,6] $\phi_i(f, x)$, under the interpretation provided by Janzing et al.[4], is equivalent to $\alpha_i(x)$ formalized in equation (11).